\newif\ifarxiv
\arxivtrue % \arxivtrue / \arxivfalse
\documentclass[a4paper,10pt]{article}

\usepackage[margin=2.5cm]{geometry}
\usepackage{caption}
\usepackage{graphicx}			% For including pictures
\usepackage[colorlinks=true, urlcolor=blue, citecolor=black, linkcolor=black]{hyperref}

\usepackage{subcaption}
\usepackage{xcolor}
\usepackage{amsmath, amssymb}
\usepackage{tikz}

\newcommand{\edited}[1]{{\color{black}#1}}
\newcommand{\Doerner}[1]{{\color{black}#1}}

\newcommand{\Mea}{\color{meac}\ensuremath{\mathcal{U}}\color{black}}
\newcommand{\mea}{\color{meac}U\color{black}}
\newcommand{\C}{\color{inpc}C\color{black}}
\newcommand{\scC}{\color{inpc}\scalebox{0.7}{C}\color{black}}

\definecolor{inpc}{rgb}{.0, 0.22, 0.46}
\definecolor{dfc}{rgb}{0.57, 0.36, 0.01}
\definecolor{meac}{rgb}{0.66, 0.34, 0.1}
\definecolor{regc}{rgb}{0.82, 0.25, 0.05}
\definecolor{cc}{gray}{0.4}

\newcommand{\df}{\color{dfc}\mathcal{D}\color{black}}
\newcommand{\Inp}{\color{inpc}\mathcal{Y}\color{black}}
\newcommand{\inp}{\color{inpc}Y\color{black}}
\newcommand{\B}{\color{inpc}B\color{black}}
\newcommand{\Reg}{\color{regc}\mathcal{R}\color{black}}
\newcommand{\Cra}{\color{inpc}\mathcal{C}\color{black}}

\title{Allure of Craquelure: A Variational-Generative Approach to Crack Detection in Paintings}

\author{%
    Laura Paul\textsuperscript{1,2}, 
    Holger Rauhut\textsuperscript{1,2}, 
    Martin Burger\textsuperscript{3,4}, 
    Samira Kabri\textsuperscript{3}, 
    Tim Roith\textsuperscript{2,5}%
    \vspace{10pt}\\
    \textsuperscript{1}Dept. of Mathematics, LMU Munich, Germany\\
    \textsuperscript{2}Munich Center for Machine Learning, Germany\\
    \textsuperscript{3}Helmholtz Imaging, Deutsches Elektronen-Synchrotron DESY, Germany\\
    \textsuperscript{4}Fachbereich Mathematik, University of Hamburg, Germany\\
    \textsuperscript{5}CIT School, Technical University of Munich,Germany
}

\date{}	% Suppress any date

\begin{document}

% To do list
%\listoftodos

\ifarxiv
    \begin{figure*}[b]
        \centering
    	\includegraphics[width=0.48\linewidth]{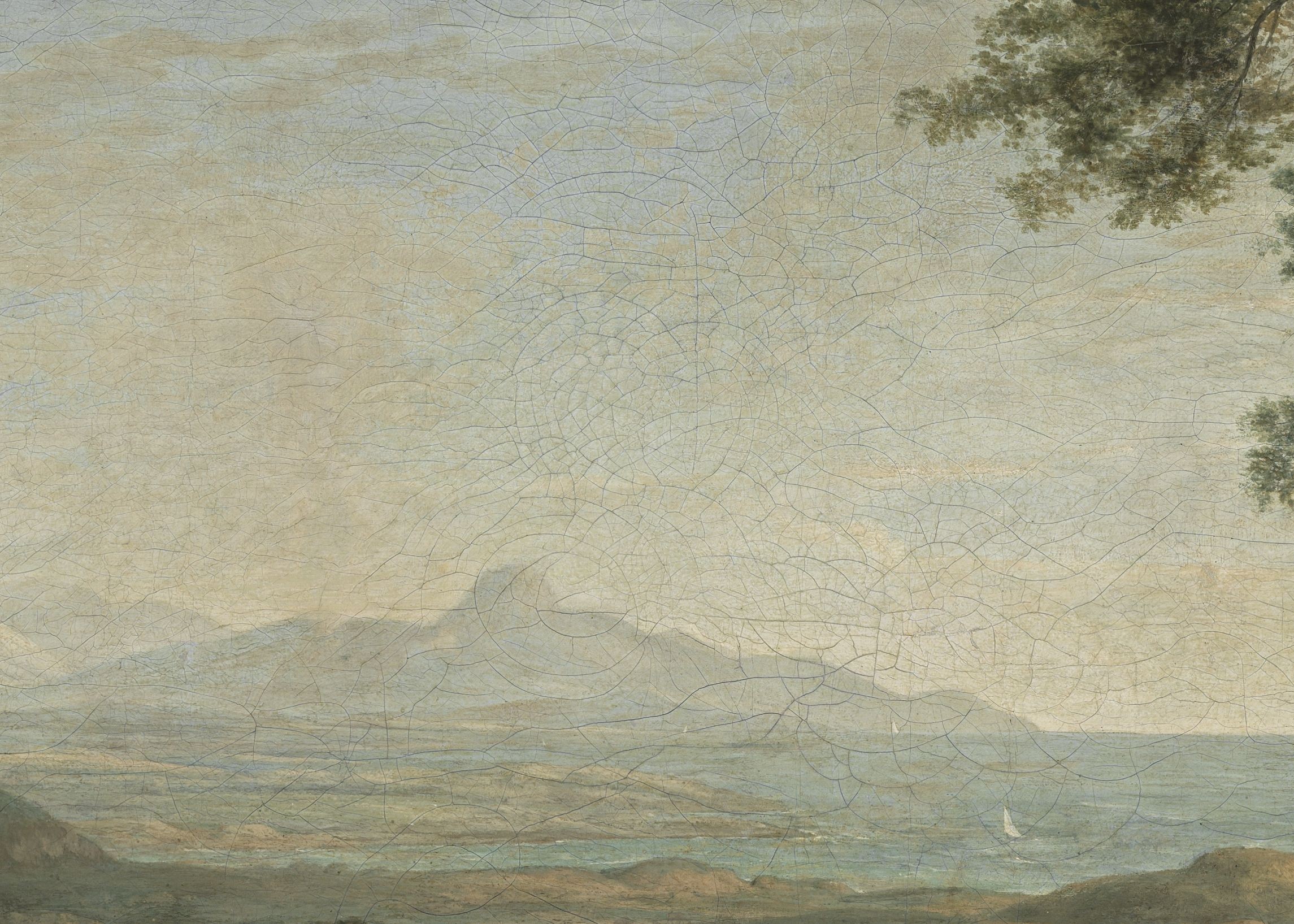}
        \hfill
    	\includegraphics[width=0.48\linewidth]{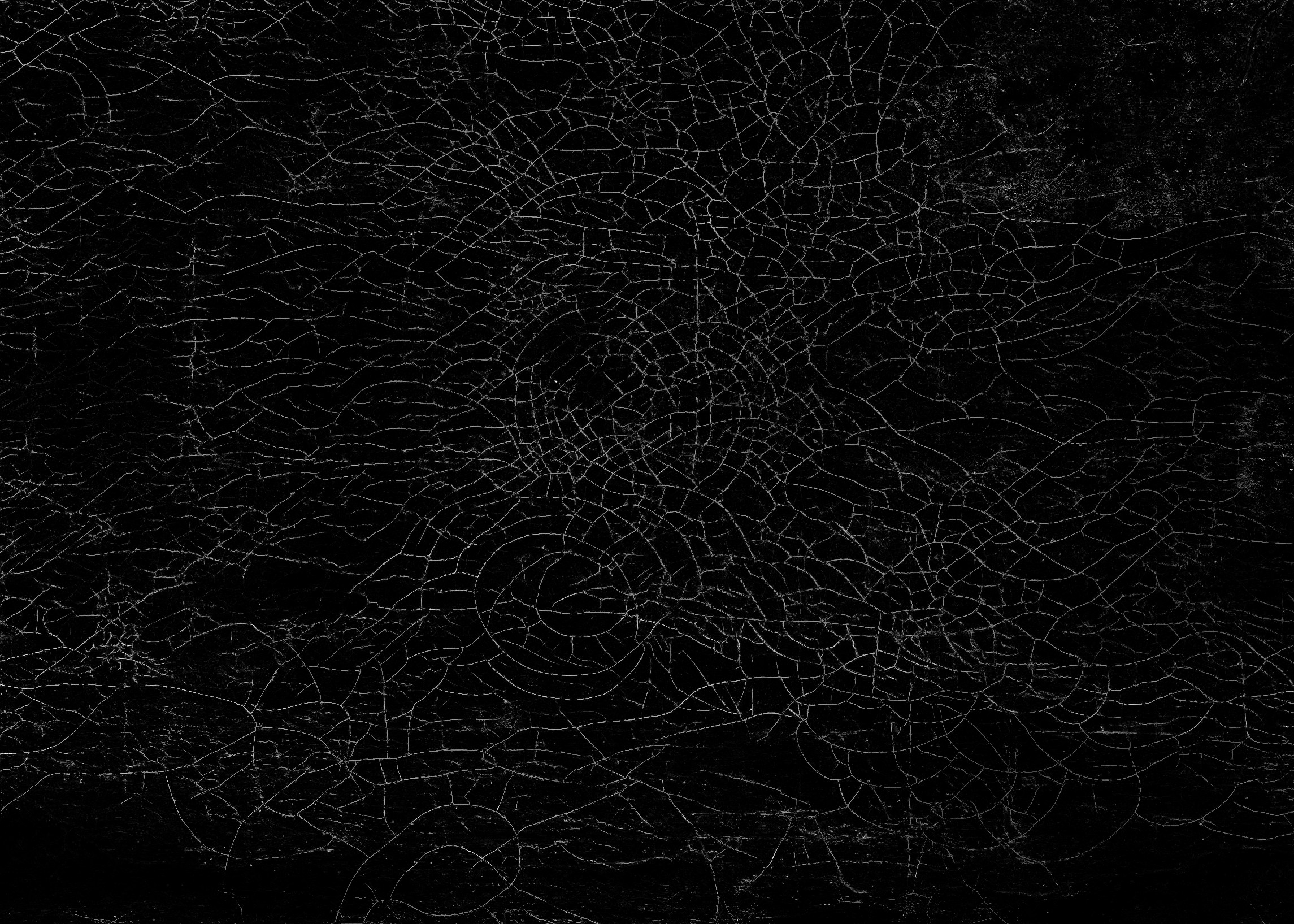}
    \caption{Qualitative example of the proposed GenAT crack detection framework. Left: cropped detail of Hagar und Ismael in der Wüste by Claude Lorrain (Claude Gellée), 1668, courtesy Bayerische Staatsgemäldesammlungen – Alte Pinakothek München. Right: detected cracks.}
    \label{fig:teaser}
    \end{figure*}
\else
    \teaser{
    \centering
    \mbox{} \hfill 
    	\includegraphics[width=0.43\linewidth]{figures/LORAIN_Wueste.jpg}
    \hfill
    	\includegraphics[width=0.43\linewidth]{figures/LORAIN_Wueste_FT4_ufdcfix_gT_lab_clahe_clahe0p010_kernel192_2mu_crack_soft_before_threshold_iter800.jpg}
    \hfill \mbox{}
    \caption{%
    \label{fig:teaser} Qualitative example of the proposed GenAT crack detection framework. Left: cropped detail of Hagar und Ismael in der Wüste by Claude Lorrain (Claude Gellée), 1668, courtesy Bayerische Staatsgemäldesammlungen – Alte Pinakothek München. Right: detected cracks.%
    }
    }
\fi

\maketitle

%-------------------------------------------------------------------------
\begin{abstract}
\ifarxiv \noindent \fi%
Recent advances in imaging technologies, deep learning and numerical performance have enabled non-invasive detailed analysis of artworks, supporting their documentation and conservation. In particular, automated detection of craquelure in digitized paintings is crucial for assessing degradation and guiding restoration, yet remains challenging due to the possibly complex scenery and the visual similarity between cracks and crack-like artistic features such as brush strokes or hair. We propose a hybrid approach that models crack detection as an inverse problem, decomposing an observed image into a crack-free painting and a crack component. A deep generative model is employed as powerful prior for the underlying artwork, while crack structures are captured using a Mumford--Shah-type variational functional together with a crack prior. Joint optimization yields a pixel-level map of crack localizations in the painting.

\smallskip

\edited{\ifarxiv \noindent \fi \textbf{Keywords:} Crack Detection, 
%Cultural Heritage Imaging,
Variational Models, Painting Analysis,
Ambrosio--Tortorelli Functional, Generative Image Priors}
%-------------------------------------------------------------------------
%  ACM CCS 1998
%  (see https://www.acm.org/publications/computing-classification-system/1998)
% \begin{classification} % according to https://www.acm.org/publications/computing-classification-system/1998
% \CCScat{Computer Graphics}{I.3.3}{Picture/Image Generation}{Line and curve generation}
% \end{classification}
%-------------------------------------------------------------------------
%  ACM CCS 2012, https://www.acm.org/publications/class-2012)
\ifarxiv
\else
\begin{CCSXML}
<ccs2012>
   <concept>
       <concept_id>10010405.10010469.10010470</concept_id>
       <concept_desc>Applied computing~Fine arts</concept_desc>
       <concept_significance>500</concept_significance>
       </concept>
   <concept>
       <concept_id>10002950.10003714.10003716</concept_id>
       <concept_desc>Mathematics of computing~Mathematical optimization</concept_desc>
       <concept_significance>500</concept_significance>
       </concept>
   <concept>
       <concept_id>10010147.10010178.10010224.10010245.10010247</concept_id>
       <concept_desc>Computing methodologies~Image segmentation</concept_desc>
       <concept_significance>500</concept_significance>
       </concept>
 </ccs2012>
\end{CCSXML}

\ccsdesc[500]{Applied computing~Fine arts}
\ccsdesc[500]{Mathematics of computing~Mathematical optimization}
\ccsdesc[500]{Computing methodologies~Image segmentation}

\printccsdesc
\fi
\end{abstract}  
%-------------------------------------------------------------------------
\section{Introduction}
\begin{figure}[tbp]
\centering
\ifarxiv
\includegraphics[width=0.48\linewidth]{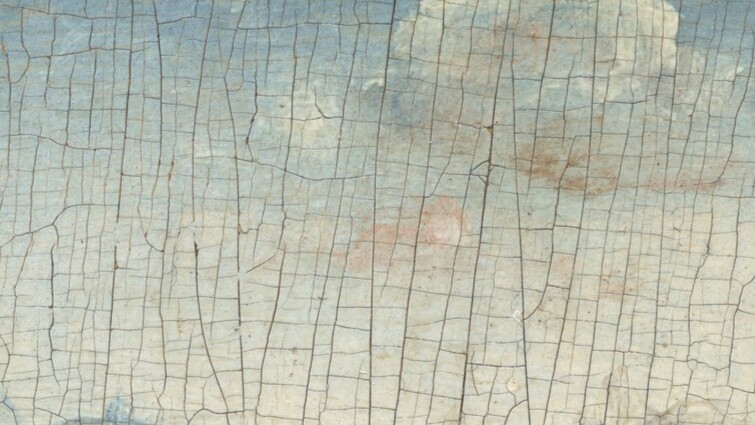}
\else
\includegraphics[width=0.49\linewidth]{figures/dark_cracks_1.jpg}
\fi
\hfill
\ifarxiv
\includegraphics[width=0.48\linewidth]{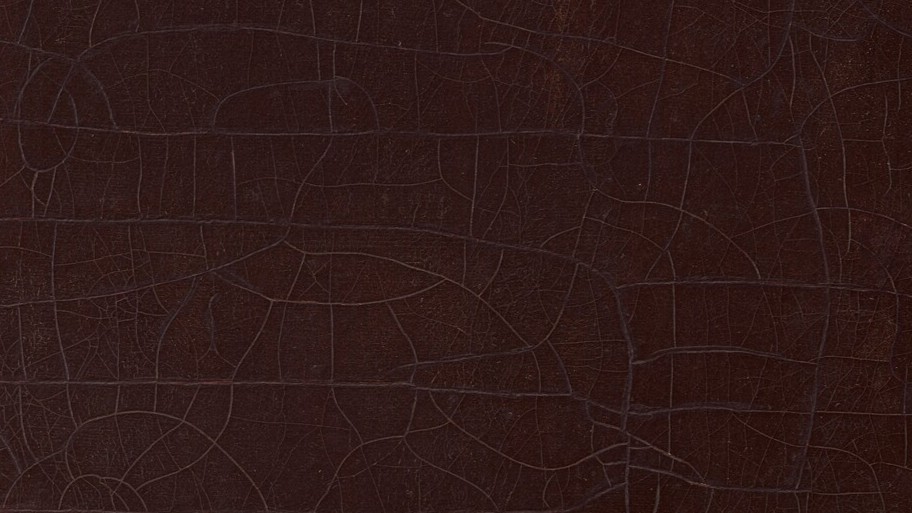}
\else
\includegraphics[width=0.49\linewidth]{figures/bright_cracks_1.jpg}
\fi
\caption{\label{fig:cracktypes} Different crack types, image courtesy of the National Gallery of Art (NGA). Left: dark appearing cracks. Right: bright appearing cracks.}
\end{figure}

Advances in imaging technologies, such as high-resolution photography, combined with recent breakthroughs in deep learning, have opened new avenues for the investigation, documentation, and conservation of artworks. These methods enable detailed analysis of the painting's internal and surface structure in a non-invasive way, without the need for physical intervention, preserving the artwork's integrity while offering deep insight into its material composition and condition. A broad overview of the application of computational and mathematical tools to art investigation is provided by Sober et al. \cite{sober2022revealing}, emphasizing the relevance and the various prospects of interdisciplinary projects. A particularly important application of mathematical and computational techniques lies in the automated detection of \textit{craquelure} patterns in digitized paintings, since cracks are among the most common forms of degradation found in aged artworks and carry significant information about the painting, such as its age, environmental history, and structural stability. Craquelure appears as a pattern of fine cracks in the paint surface, which can appear either dark or bright depending on the underlying paint and lighting conditions. Examples can be observed in Figure \ref{fig:cracktypes}.
Detecting and analyzing the cracks accurately is essential for conservators in planning and executing restoration measures. Moreover, automatic crack detection is highly relevant in the context of Cultural Heritage (CH) and museum digitization workflows, since large collections of high-resolution paintings need to be documented, monitored, and analyzed in a scalable and non-invasive manner. Reliable crack localization can support conservators in condition assessment, restoration planning, long-term damage monitoring, and digital archiving, while reducing the need for time-consuming manual inspection \cite{dulecha2019crack, sober2022revealing}. One huge difficulty though is to develop methods that can distinguish cracks from visually similar elements such as brush strokes, canvas structure or painted hair, that are part of the original artistic process and appearance. Crack detection in digitized artworks using classic mathematical image processing methods dates back to the 1990s. Early works used morphological filters like the top-hat transformation in order to display cracks \cite{giakoumis_digital_1998, abas2003}. Similar methods were investigated by Zana and Klein for biomedical imaging in order to display vessel-like structures \cite{zana_segmentation_2001}. After that, many works proposed morphological filtering as a preprocessing step combined with machine learning based postprocessing methods in order to distinguish cracks from visually similar structures such as brush strokes or hair that had previously been falsely identified as cracks \cite{giakoumis_digital_2006, spagnolo_virtual_2010, ruvzic_2011, cornelis_crack_2012, sizyakin_deep_2018, rucoba-calderon_crack_2022}. Among these approaches, the methods of Giakoumis et al. \cite{giakoumis_digital_2006}, Spagnolo and Somma \cite{spagnolo_virtual_2010}, Ružić et al. \cite{ruvzic_2011}, and Rucoba-Calderón et al. \cite{rucoba-calderon_crack_2022} are largely unsupervised, as they rely on handcrafted image processing, restoration techniques, or data-driven models that do not require pixel-wise crack annotations for training. In contrast, Cornelis et al. \cite{cornelis_crack_2012} employ a semi-supervised postprocessing step, while Sizyakin et al. \cite{sizyakin_deep_2018} use supervised learning based on annotated crack examples.  Furthermore, the Bayesian Crack Tracing Framework (BCTF) proposed by Cornelis et al. \cite{cornelis_2013, cornelis_2015} combines morphological preprocessing and handcrafted feature extraction with probabilistic Bayesian crack tracing in order to suppress false detections caused by painting structures. The method is semi-supervised in the sense that statistical models are learned from annotated crack examples, while the final crack tracing is performed through a probabilistic inference procedure rather than direct pixel-wise classification. More recent works investigate supervised convolutional neural network approaches for crack detection in digitized painting patches \cite{dulecha2019crack, sizyakin_crack_2020}. The most recent work in the field of art investigation considers a supervised segmentation convolutional autoencoder (U-Net) for crack detection \cite{sizyakin_deep_2022}. However, the authors report a high sensitivity of the method to the accuracy and completeness of manual crack annotations, so that ideally all cracks in the training dataset should be labeled. Several of the mentioned approaches in the CH literature additionally rely on multi-modal or multi-acquisition imaging data, such as X-ray, infrared, or multi-light imaging, in order to better separate cracks from the underlying painting structure \cite{cornelis_2015, sizyakin_deep_2018, dulecha2019crack, sizyakin_crack_2020, sizyakin_deep_2022}. Beyond the CH setting, crack detection has also been extensively studied for structural damage analysis in roads and pavements, see for example \cite{zhang2016road, DeepCrack}. In this context, modern deep learning based segmentation approaches have achieved strong performance due to the availability of large annotated datasets and the comparatively homogeneous visual appearance of road surfaces. Nevertheless, images in this setting are far less complex than artworks, and various datasets with ground-truth annotations are readily available for training and evaluation, which are lacking in the arts context.

In the following, we propose a novel \edited{hybrid variational-generative} approach to crack detection in digitized paintings, \edited{formulating the task as an inverse problem. The method combines a deep generative model with a Mumford--Shah-type functional and an optional learned crack prior to separate cracks from the underlying painting. The general framework accommodates crack priors and parameter-selection strategies with different supervision levels and can
therefore be instantiated in a fully unsupervised manner. In the implementation presented here, the chosen prior is fine-tuned and the regularization parameters are selected using labeled synthetic data. We therefore characterize the evaluated method as weakly supervised.} In contrast to several existing CH approaches, our method operates solely on standard RGB digitized images and does not require multi-modal acquisition data such as infrared or X-ray imaging. Moreover, our method neither requires image preprocessing nor postprocessing beyond optional contrast enhancement and a thresholding step. It produces pixel-level binary maps localizing cracks in digitized artworks; Figure \ref{fig:teaser} provides a representative example of the results.

\section{Crack Detection in Digitized Paintings as Inverse Problem}
\label{sec:theory}
In this work, the task of crack detection in images will be considered an \textit{inverse problem}, thus we give a brief introduction on that topic based on \cite{arridge2019solving} before formulating the problem in our precise setting. An inverse problem consists of recovering an unknown signal or parameter 
$\inp \in \Inp$ from measurements $\mea \in \Mea$ generated by a forward operator $T : \Inp \to \Mea$, potentially corrupted by noise $n$, that is 
\[
\mea = T(\inp) + n(\inp).
\]
Inverse problems are said to be \textit{ill-posed}, if the solution is sensitive to variations in data. More precisely, that is when at least one of the following fails: a solution may not exist for all measurements $\mea$ (existence), multiple $\inp$ may explain the same $\mea$ (uniqueness), or small perturbations in $\mea$ may cause large deviations in $\inp$ (stability). Such difficulties are very common, so regularization strategies are required to overcome them. In many applications, the noise can be assumed to be independent of the unknown input $\inp$, which yields $n(\inp) \equiv n$. In our specific case however, this simplification does not apply, since the observed corruption is caused by cracks whose appearance depends on the underlying painting content. More precisely, we consider $\Inp = \Mea \times \Cra$ and inputs $\inp = (\B, \C)$, where $\B$ models the crack-free image part, and $\C$ models the crack pattern. The forward operator $T$ is a projection of $(\B,\C)$ onto the crack-free image $B$ while the noise is parametrized by the crack pattern $\C$. Consequently, the forward problem reads  
\begin{align}
\label{eq:hierarchicalnoiseforwardprob}
    \mea = \B + n(\C)
\end{align}
and the inverse problem consists of retrieving $(\B,\C)$ from the measurement $\mea$. The crack pattern $\C$ describes the spatial distribution of cracks, while the corresponding image corruption is represented by the term $n(\C)$. In the context of digital crack detection, $\mea \in \Mea = \mathbb{R}^{h \times w \times 3}$ %
%Continuous: $U: \Omega \to \mathbb{R}^3, \Mea=L^2(\Omega;\mathbb{R}^3)$
is the input RGB image, that is the digitized painting, and the goal is to recover the crack probability map $\C \in \Cra =[0,1]^{h \times w}$ %
%Continuous: $\C:\Omega \to [0,1]$
as well as the crack-free painting part $\B \in \mathbb{R}^{h \times w \times 3}$ from it while the forward operator $T$ maps $(\B,\C) \in \Inp = \mathbb{R}^{h\times w \times 3} \times [0,1]^{h \times w}$ %
%Continuous: $\mathcal{Y}= (L^2(\Omega;\mathbb{R}^3) \times L^2(\Omega))$
to the observed image. In this paper, we consider the following variational regularization approach to reconstruct $\B$ and $\C$:
\[
\min_{\B, \scC} \; 
\underbrace{\df(\B,\C, \mea)}_{\text{data fidelity}} 
+ \underbrace{\Reg_{\text{Painting}}(\B, \C)}_{\text{painting prior}} 
+ \underbrace{\Reg_{\text{Crack}}(\C)}_{\text{crack prior}}.
\]
\begin{itemize}
\item \color{dfc}{\textit{Data fidelity}}\color{black}: The function $\df:\Mea\times\Cra\times\Mea\to[0,\infty)$ compares the similarity of the crack-free image $\B$ to the measurement $\mea$, taking into account the crack pattern $\C$. Intuitively, the data fidelity measures how likely it is that the measurement $\mea$ was created by the quantity $(\B,\C)$ in the forward problem \eqref{eq:hierarchicalnoiseforwardprob}.
\item \color{inpc}{\textit{Painting prior}}\color{black}: The term \emph{prior} originates from the Bayesian viewpoint, where before any measurement is taken we have a prior belief about the appearance of paintings $\B$. In the regularization formulation $\Reg_{\text{Painting}}(\B,\C)$ has small values for images $\B$ with typical features and high values otherwise. The painting prior may depend on the crack pattern $\C$ since it should not assess the painting in cracked regions.
\item \color{cc}{\textit{Crack prior}}\color{black}: Similarly, the crack prior incorporates our prior belief on the shape of cracks $\C$.
\end{itemize}
In the following, we recall a classical mathematical framework for modeling image discontinuities, namely the \textit{Mumford--Shah functional}.

\subsection{Mumford--Shah formulation}
For modeling purposes, in contrast to the previous discrete treatment, from now on we adopt a continuous formulation. In the numerical implementation, the corresponding functionals are discretized on the pixel grid. The Mumford--Shah (MS) functional \cite{Mumford-Shah}, originally introduced for image segmentation to approximate an image by a piecewise smooth function separated by a set of discontinuities, provides a natural variational framework for modeling cracks, as the discontinuity set $K$ corresponds to the crack locations from which the crack pattern $\C$ is inferred. The MS regularizer is defined as %
\[
\Reg_{\text{MS}}(\B, K) 
= \int_{\Omega \setminus K} |\nabla \B|^2 \, dx + \mu ~\mathcal{H}^1(K),
\]%
where $\Omega \subset \mathbb{R}^2$ is the image domain, $\B$ is supposed to be in the Sobolev space $H^1(\Omega\setminus K)$, $K \subset \Omega$ denotes the set of discontinuities (edges), $\mathcal{H}^1(K)$ its Hausdorff measure (length), and $\mu>0$ balances the edge length. This functional encourages $\B$ to be smooth away from edges while allowing sharp discontinuities at crack locations. In practice, the MS functional is approximated using the \textit{Ambrosio--Tortorelli} (AT) relaxation \cite{Ambrosio-Tortorelli}, introducing an auxiliary edge indicator function $v: \Omega \to [0,1]$:%
\[
\Reg_{\text{AT}}(\B, v) = \underbrace{\int_\Omega v^2 |\nabla \B|^2 dx}_{\text{painting regularity}} + ~\mu \underbrace{\int_\Omega \Big( \varepsilon |\nabla v|^2 + \frac{(v-1)^2}{4\varepsilon} \Big) dx}_{\text{crack regularity}},
\]%
where $\varepsilon > 0$ controls the smooth transition around cracks, and $\mu$ weights the crack length. In this formulation, $v \approx 0$ at crack locations and $v \approx 1$ elsewhere, so it highlights cracks while allowing $\B$ to be smooth between cracks. To highlight the distinction between the regularization performed on the painting and the one performed on the crack pattern, we denote the two terms separately by
\begin{gather}
\Reg_{\text{PReg}}(\B, v) := \int_\Omega v^2 |\nabla \B|^2 dx, \\
\Reg_{\text{CReg}}^\varepsilon(v) := \int_\Omega \Big( \varepsilon |\nabla v|^2 + \frac{(v-1)^2}{4\varepsilon} \Big) dx. \notag
\label{GradientTerm}
\end{gather}
We then choose the data-fidelity term for fixed $\sigma^2 > 0$ as
\[
   \df(\B, v, \mea) := \frac{1}{\sigma^2} \|v \big(\mea - \B\big)\|_2^2 = \frac{1}{\sigma^2} \int_\Omega v(x)^2 |\mea(x)-\B(x)|^2 dx,
\] %
so that the observed image is effectively compared to the reconstructed background only in crack-free regions, since $v$ is close to one away from cracks and close to zero at crack locations. Consequently, discrepancies between $\mea$ and $\B$ are penalized strongly in smooth painting regions, while deviations at cracks are only weakly penalized and can therefore be absorbed by the crack component. 
This kind of fidelity term can be interpreted as arising from a weighted Gaussian noise model where cracks are viewed as a spatially varying corruption $\eta$ of the observed image $\mea$. Formally, one may think of $\mea = \B + \eta$, where the local variance depends on $v$. At the discrete image level, this corresponds to the heteroscedastic Gaussian model
\[
    \mea_i = \B_i + \eta_i,
    \quad \eta_i \sim \mathcal{N} \Bigg( 0, \frac{\sigma^2}{v_i^2} \Bigg),
\]
independently over pixels $i$. % 
%\todo[inline]{TR: I'm really not sure if the noise model can be easily defined in infinite dimensions. I would prefer to only keep the discrete level here. Ok better be safe.}
% \todo[inline]{TR: is the below well-defined if $v$ is a function?\\ LP: No, i changed it.}
% \todo[inline]{LP: Oder so: At the discrete image level, this corresponds to the heteroscedastic Gaussian
% model
% \[
%     \mea_i = \B_i + \eta_i,
%     \quad \eta_i \sim \mathcal{N} \Bigg( 0, \frac{\sigma^2}{v_i^2} \Bigg),
% \]
% independently over pixels $i$. 
% %
% }
% \[
%     \operatorname{Var}(\eta(x)) \approx \frac{\sigma^2}{v(x)^2},
%     \qquad x\in\Omega.
% \]
Thus, the variance becomes large in regions where $v(x) \approx 0$, meaning that observations near cracks are treated as unreliable, while in crack-free regions where $v(x) \approx 1$, the variance is approximately $\sigma^2$ and the observed image is trusted.

To incorporate prior knowledge of plausible crack patterns, we employ a learned crack estimator $P$. This further helps to reduce false positives while preserving fine crack structures. More precisely, we introduce another regularization functional%
\[
\Reg_{\mathrm{CP}}(\mea, v) := \int_\Omega \big( v - P(\mea) \big)^2 ~dx.
\]
In this formulation, $P(\mea):\Omega\to [0,1]$ gives an estimate of the crack pattern present in the input painting $\mea$, where values close to zero indicate cracks and values close to one correspond to crack-free regions. This encourages the predicted crack map to be consistent with the prediction obtained by $P$. \edited{The supervision level of the resulting method depends in particular on the construction of $P$. We refer to Section~\ref{sec:experiments} for details on its implementation}.
% \[
% \Reg_{\mathrm{CP}}(\mea^\prime, v) = \int_\Omega \big( v - P(\mea^\prime) \big)^2 ~dx.
% \]%
% \todo[inline]{TR: maybe instead
% \[
% \Reg_{\mathrm{CP}}(\mea^\prime, v) = \int_\Omega \big( v \cdot  (1-P(\mea^\prime)) \big)^2 ~dx.
% \]%
% }
% In this formulation, $P(\mea^\prime):\Omega\to [0,1]$ gives an estimate of the crack pattern present in the input painting $\mea^\prime$, where values close to zero indicate cracks and values close to one correspond to crack-free regions. This encourages the predicted crack map to be consistent with the prediction obtained by $P$. We additionally optimize over the input $\mea'$ and initialize it with the original input image $\mea$. We refer to the section on Numerical Experiments for details on the choice of $P$.

\subsection{Introducing a generative background prior} To further constrain the space of admissible crack-free paintings and to mitigate the ill-posedness of the decomposition, we model the background component $\B$ using a deep generative prior. That is, we assume that $\B$ lies in the range of a generator $G : \mathcal{Z} \to \mathcal{X}$, mapping from a  low-dimensional latent space \(\mathcal{Z} \subset \mathbb{R}^d\) into the space of RGB images \(\mathcal{X}\). Restricting $\B$ to the generator range effectively constrains the solution to a learned manifold of plausible natural images, which has been shown to substantially reduce the ambiguity of ill-posed inverse problems \cite{bora2017compressed, dimakis2022deep, duff2024regularising}. In our setup, the latent variable $z$ is optimized jointly with the crack indicator within the proposed variational formulation, allowing the generative prior to act as an implicit regularizer for the background reconstruction.

\subsection{The full optimization problem} 
Let $G(z)$ denote the generative reconstruction of the crack-free painting and $z$ its latent code. Combining the generative painting prior, the AT crack prior, and the additional crack prior, we define
\ifarxiv
\begin{align*}
\Reg(G(z), v, P(\mea)) 
:= \lambda_{\text{PReg}}~\Reg_{\mathrm{PReg}}(G(z), v)
+ \lambda_{\text{CReg}}~\Reg_{\mathrm{CReg}}^\varepsilon(v)
+ \lambda_{\mathrm{CP}}~\Reg_{\mathrm{CP}}(P(\mea), v), \notag
\end{align*}
\else
\begin{align*}
&\Reg(G(z), v, P(\mea)) \\
& := \lambda_{\text{PReg}}~\Reg_{\mathrm{PReg}}(G(z), v)
+ \lambda_{\text{CReg}}~\Reg_{\mathrm{CReg}}^\varepsilon(v)
+ \lambda_{\mathrm{CP}}~\Reg_{\mathrm{CP}}(P(\mea), v), \notag
\end{align*}
\fi
where the hyperparameters $\lambda_{\text{PReg}}, \lambda_{\text{CReg}}, \lambda_{\text{CP}} \geq 0$ determine the influence of the different regularization terms. The resulting full optimization problem is then given by
\begin{align}
\label{functional}
\min_{z, v}
%~\big\| v (\mea - G(z)\color{black}) \big\|_2^2 
~\df(G(z),v,\mea) ~+~ \Reg(G(z), v, P(\mea)).
\end{align}
% \begin{gather*}
% \min_{z, v, \mea'} ~\big\| v \odot (\mea - G(z)\color{black}) \big\|_2^2 \\
% + \lambda_{\text{PReg}}~\Reg_{\mathrm{PReg}}(G(z), v) + \lambda_{\text{CReg}}~\Reg_{\mathrm{CReg}}^\varepsilon(v) 
% + \lambda_{\mathrm{CP}} ~\Reg_{\mathrm{CP}}(P(\mea'), v),
% \end{gather*}
We refer to the proposed approach as \emph{GenAT}. After optimization, the predicted pixel-level crack map is obtained as
\[
\C_{\text{pred}}(x) = 1 - v(x), \quad x \in \Omega,%
\]
so that brighter pixels correspond to probable crack locations. 
In the classical Ambrosio--Tortorelli formulation, the mixed gradient regularization term $\Reg_{\mathrm{PReg}}(G(z),v)$ penalizes image gradients in regions where $v$ is close to one and therefore promotes piecewise smooth reconstructions away from cracks. However, in the context of paintings, strong gradients naturally occur due to semantic image content such as brush strokes, object boundaries, facial features, or hair structures. Enforcing the mixed gradient term too strongly may therefore incorrectly attribute such structures to cracks or overly smooth relevant artistic details in the reconstructed crack-free image. For this reason, we consider and experimentally evaluate both variants of the model, namely one including the mixed gradient regularization term and one without it. This allows us to investigate whether relaxing the classical AT assumption of piecewise-smooth image regions leads to improved crack detection performance in complex artwork scenes. To distinguish the two variants considered in this work, we write GenAT $\Reg_{\mathrm{PReg}}$ whenever the mixed gradient regularization term $\Reg_{\mathrm{PReg}}(G(z),v)$ is included in the optimization problem. Otherwise, we simply write GenAT.
%
%%%%%%%%%%%%%%%%%%%%%%%%%%%%%%%%%%%%%%%
\section{Numerical experiments}
\label{sec:experiments}
In this section, we describe the implementation of the proposed approach and report crack detection results obtained on a synthetic dataset as well as on real-world cracked paintings. While the variational crack model is formulated in the continuous setting for modeling purposes, the training objectives considered in this section are implemented on discrete image grids and therefore written in a discrete way. All experiments were performed on an NVIDIA A100 GPU. For both the representation of intact regions and the improved regularization of crack locations, we make use of existing machine learning models:
\begin{itemize}
    \item For the generator $G$ we use a fine-tuned version of the VQGAN model \cite{vqgan}, pretrained on the ImageNet dataset \cite{deng2009imagenet}. VQGAN combines vector-quantized variational autoencoders with adversarial training, yielding a compact discrete latent representation that enables high-fidelity image synthesis while preserving local texture and structural consistency. This makes it well suited for modeling crack-free painting backgrounds, as it captures realistic material texture and spatial coherence without the blurring artifacts typical of pixel-space autoencoders.
    \item As crack prior $P$ we use a fine-tuned version of the DeepCrack model proposed by Zou et al. \cite{DeepCrack}. DeepCrack is a convolutional neural network that was trained on pavement images and outputs per-pixel crack probabilities from an input image. 
\end{itemize}
\edited{The evaluated implementation is weakly supervised, since DeepCrack is fine-tuned using pixel-wise synthetic crack masks, and regularization
parameters are selected on a labeled synthetic validation set. The code for reproducing the experiments is publicly available at \url{https://github.com/Lauringel/GenAT}.}

\subsection{Datasets} 
\begin{figure}[tbp]
\centering
\begin{minipage}[b]{%
\ifarxiv%
0.18\textwidth
\else%
0.258\linewidth
\fi%
}
\includegraphics[width=\linewidth]{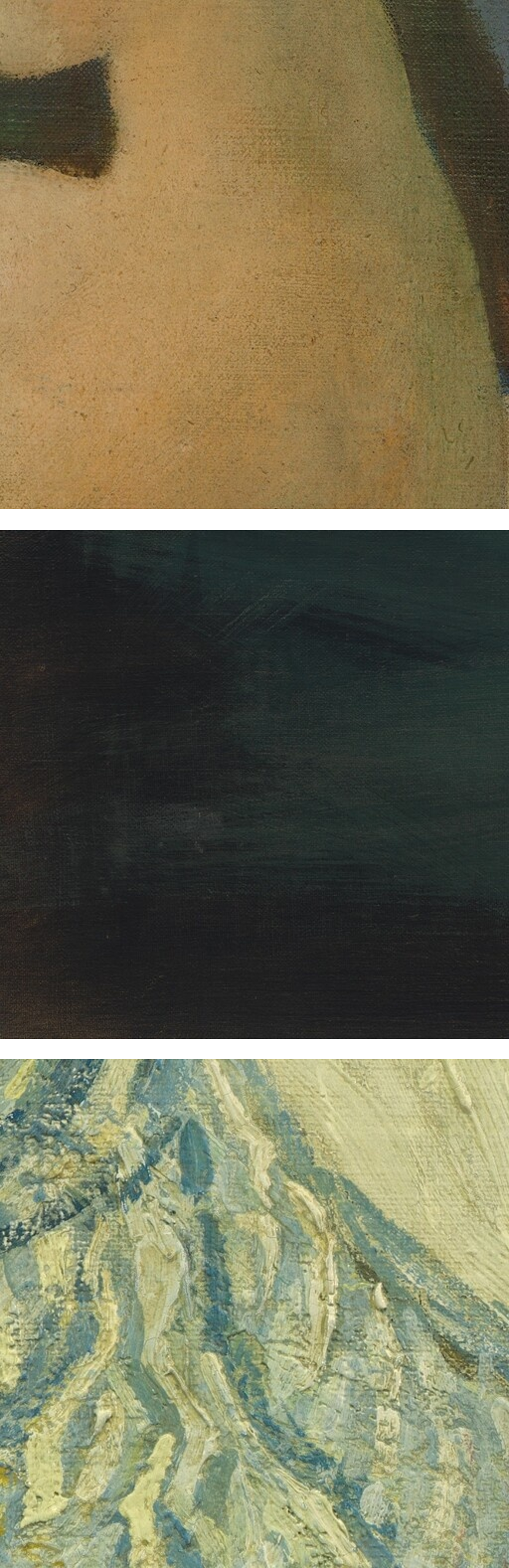}
        	\subcaption{} % Add subcaption text if desired, or use \subcaption* to suppress (a), (b), etc. labels
        	\label{fig:dsa}
\end{minipage}
\ifarxiv
\quad
\else
\hfill %add desired spacing between images, e. g., ~, \quad, \qquad, \hfill etc.
\fi
\begin{minipage}[b]{%
\ifarxiv%
0.18\textwidth
\else%
0.258\linewidth
\fi%
}
	\includegraphics[width=\linewidth]{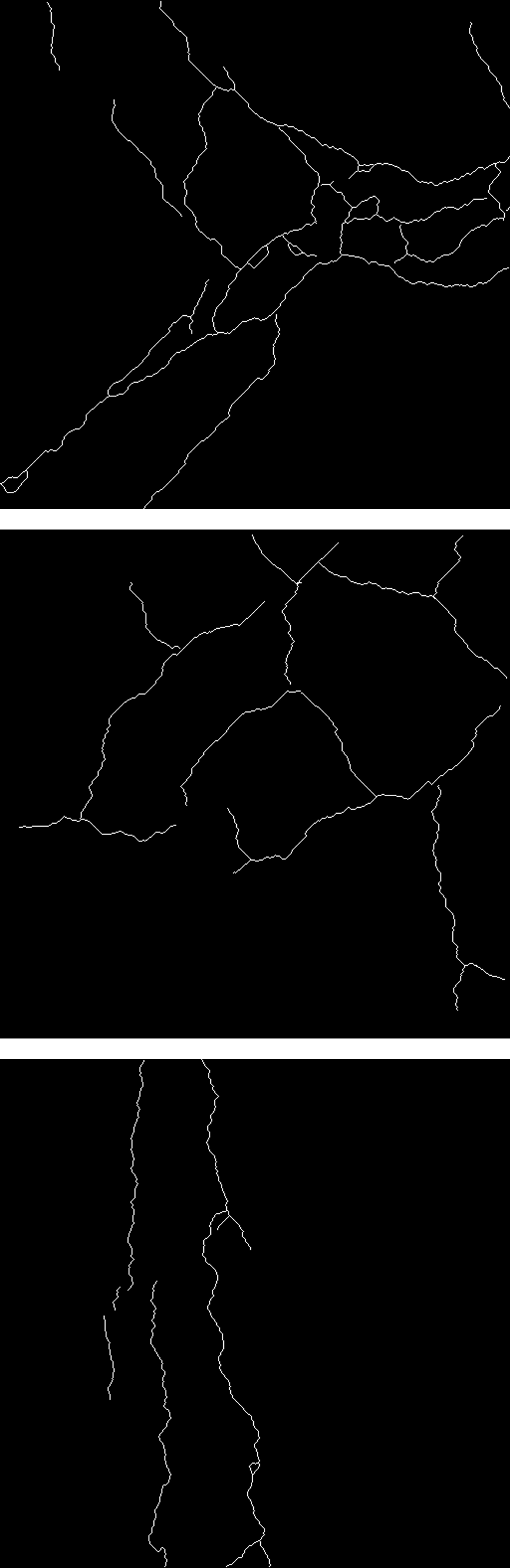}
        	\subcaption{} % Add subcaption text if desired, or use \subcaption* to suppress (a), (b), etc. labels
        	\label{fig:dsb}
\end{minipage}
\ifarxiv
\quad
\else
\hfill %add desired spacing between images, e. g., ~, \quad, \qquad, \hfill etc.
\fi
\begin{minipage}[b]{%
\ifarxiv%
0.18\textwidth
\else%
0.258\linewidth
\fi%
} %0.28
	\includegraphics[width=\linewidth]{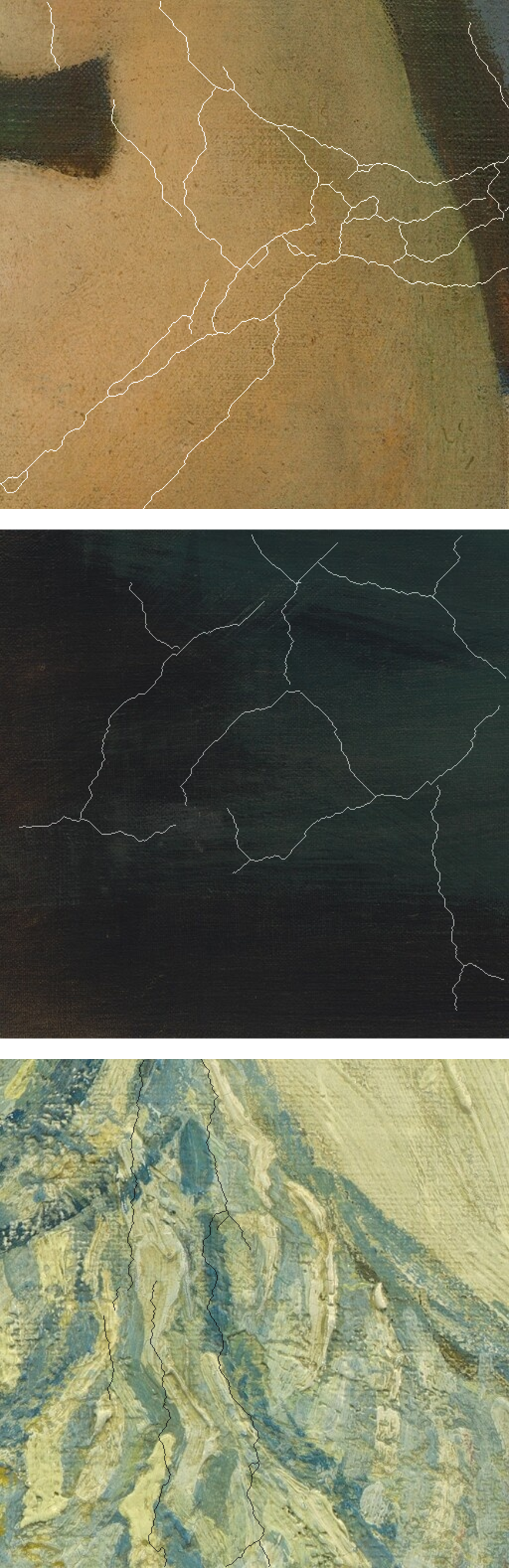}
        	\subcaption{} % Add subcaption text if desired, or use \subcaption* to suppress (a), (b), etc. labels
        	\label{fig:dsc}
\end{minipage}
\caption{Illustration of the construction of the synthetic dataset: (a) real painting patches without visible cracks, courtesy NGA, (b) real crack patterns from \cite{DeepCrack}, and (c) created synthetic cracked painting patches.}
\label{fig:dataset}
\end{figure}

A major challenge in crack detection for artworks is the lack of pixel-wise ground-truth annotations. To quantitatively evaluate our approach and overcome this limitation, we construct a synthetic dataset of cracked paintings. As a basis, we use a self-curated dataset of approximately $8100$ crack-free painting patches of size $512 \times 512$ pixels. The source images are digitized paintings obtained from the National Gallery of Art (NGA), the Slovak National Gallery, and Wikimedia Commons. \edited{The source paintings span different artistic styles and historical periods.} A selection can be seen in Figure \ref{fig:dsa}. The synthetic dataset of approximately $1700$ images with cracks is then generated by overlaying realistic crack patterns onto a subset of the crack-free painting patches. To obtain plausible crack geometries, we leverage publicly available pavement crack datasets, specifically CrackLS315 and CRKWH100 \cite{DeepCrack}, using their binary ground-truth masks (see Figure \ref{fig:dsb}). The masks are blended into the painting images via an alpha-compositing model, where the crack appearance (dark or bright) is adaptively determined based on the average luminance of the underlying painting patch, as real craquelure can appear either bright or dark. The resulting images simulate realistic crack patterns with varying contrast and strength while preserving the original painting structure. Examples can be observed in Figure \ref{fig:dsc} and Figure \ref{fig:comp_a}.

\subsection{Fine-tuning of the machine learning models} In our experiments, the VQGAN model is not used in its original pretrained form. Instead, it is fine-tuned in an unsupervised manner on the self-curated dataset of around 8100 crack-free images, meaning that no paired ground-truth reconstructions or pixel-level annotations are available during training. This setting is particularly challenging, since the model must adapt to the visual characteristics of artworks solely from the image data itself while preserving fine structural details and avoiding the generation of artificial crack-like artifacts. Fine-tuning is performed by unfreezing the last two encoder blocks, while keeping the codebook fixed and freezing the remaining encoder layers. The decoder and post-quantization convolution layers are trained to adapt the generator to the visual statistics of artworks. During fine-tuning, the generator is optimized using a combination of a pixel-wise reconstruction loss and a perceptual loss. The reconstruction loss is chosen depending on the variant considered later in the AT optimization. For the model including the mixed gradient term $\Reg_{\mathrm{PReg}}(G(z),v)$, the VQGAN background prior is fine-tuned using an $\ell_1$ reconstruction loss, given by
\[
\mathcal{L}_{\mathrm{rec}}^{(1)}(G(z),\B) = \| G(z)-\B \|_1.
\]
For the model without the mixed gradient term, we instead use an $\ell_2$ reconstruction loss, that is
\[
\mathcal{L}_{\mathrm{rec}}^{(2)}(G(z),\B) = \|G(z)-\B \|_2^2.
\]
This distinction is motivated by the different role of the generator in the two variants. When the mixed gradient term is present, the AT functional already imposes additional smoothness on the reconstructed background away from cracks, so the $\ell_1$ loss is used to preserve sharper image structures during fine-tuning. In the variant without the mixed gradient term, this explicit smoothing effect is removed, and the $\ell_2$ loss provides a stronger tendency towards globally smooth and stable reconstructions. 
In addition, a \textit{perceptual loss} based on pretrained VGG features is employed,
\[
\mathcal{L}_{\mathrm{perc}}(G(z), \B)
= \sum_{l} \big\| \phi_l(G(z)) - \phi_l(\B) \big\|_1,
\]
where $\phi_l(\cdot)$ denotes the activation of the $l$-th layer. The total fine-tuning objective is then given by
\[
\mathcal{L}_{\mathrm{FT}}
= \lambda_{\mathrm{rec}} \, \mathcal{L}_{\mathrm{rec}}^{(i)}
+ \lambda_{\mathrm{perc}} \, \mathcal{L}_{\mathrm{perc}}, \quad i \in \{1,2\} .
\]
We set $\lambda_{\text{rec}} = 0.65$ and $\lambda_{\text{perc}} = 0.25$, placing stronger emphasis on the reconstruction loss to favor accurate pixel-level reconstruction of crack-free painting textures while the perceptual loss complements it to maintain semantic and structural consistency without over-smoothing fine details.

We further use a fine-tuned version of DeepCrack. The potential complex and diverse background of paintings, which may contain crack-like structures such as brush strokes or painted hair, poses a significant risk of false positive detections. Since the original DeepCrack model was trained on pavement datasets, we fine-tuned the model on a training subset of our synthetic dataset in order to obtain a reliable crack prior despite the complexity of painting scenes. More precisely, the side-outputs and fusion layer of the pretrained model were retrained in a supervised way, meaning that paired training images and corresponding pixel-level crack annotations are available to explicitly guide the learning process. The retraining uses a pixel-wise \textit{Binary Cross-Entropy} (BCE) loss to encourage accurate discrimination between crack and non-crack pixels. Given an input image $\mea$ and the corresponding binary ground-truth crack mask $M \in \{0,1\}^{h \times w}$, the BCE loss is defined as %
\ifarxiv
\begin{align*}
\mathcal{L}_{\text{BCE}}(\mea) = 
- \frac{1}{|\Omega|} \sum_{x \in \Omega} M(x) \log P(\mea)(x) + (1-M(x)) \log \big(1-P(\mea)(x)\big),
\end{align*}
\else
\begin{align*}
&\mathcal{L}_{\text{BCE}}(\mea) \\
&= - \frac{1}{|\Omega|} \sum_{x \in \Omega} M(x) \log P(\mea)(x) + (1-M(x)) \log \big(1-P(\mea)(x)\big),
\end{align*}
\fi
where $M(x) \in \{0,1\}$ denotes the ground-truth crack label at pixel
$x \in \Omega$, $P(\mea)(x) \in [0,1]$ refers to the predicted crack probability and $|\Omega|$ is the pixel count.
% \begin{align*}
% &\mathcal{L}_{\text{BCE}}(\mea') \\
% &= - \frac{1}{|\Omega|} \sum_{x \in \Omega} M(x) \log P(\mea')(x) + (1-M(x)) \log \big(1-P(\mea')(x)\big),
% \end{align*}
% where $P(\mea')(x) \in [0,1]$ refers to the predicted crack probability at pixel $x \in \Omega$.

\subsection{Solving the minimization problem} 
To minimize the variational functional \eqref{functional}, we perform gradient-based optimization over the painting latent representation $z$ and the crack indicator $v$. %, and the crack prior input image $\mea'$. 
The resulting gradients are then used in an iterative scheme to update $v$ and $z$ simultaneously until convergence. We employ the Adam optimizer \cite{kingma2014adam} with separate learning rates for each variable. 
%\ifarxiv Further details on the optimization procedure are provided in the  \hyperref[sec:Appendix]{Appendix}. \fi

\subsection{Performance measures}
We assess the crack detection performance by several standard metrics for binary classification tasks. \edited{For each test image $i$, let
\begin{itemize}
    \item true positives ($\mathrm{TP}_i$) denote correctly detected crack pixels,
    \item false positives ($\mathrm{FP}_i$) denote non-crack pixels incorrectly detected as crack pixels,
    \item false negatives ($\mathrm{FN}_i$) denote crack pixels missed by the detector.
\end{itemize}}
\edited{
%The \textit{Precision} measures how reliable the detected crack pixels are and is defined as
% \[
% \mathrm{P}_i = \frac{\mathrm{TP}_i}{\mathrm{TP}_i+\mathrm{FP}_i}.
% \]
% The \textit{Recall} measures how many true crack pixels are detected and is defined as
% \[
% \mathrm{R}_i = \frac{\mathrm{TP}_i}{\mathrm{TP}_i+\mathrm{FN}_i}.
% \]
The \textit{precision} $\mathrm{P}_i$ measures how reliable the detected crack pixels are, while the \textit{recall} $\mathrm{R}_i$ measures how many true crack pixels are detected. They are defined as 
\begin{align*}
    \mathrm{P}_i = \frac{\mathrm{TP}_i}{\mathrm{TP}_i+\mathrm{FP}_i},\quad \text{and} \quad \mathrm{R}_i = \frac{\mathrm{TP}_i}{\mathrm{TP}_i+\mathrm{FN}_i}.
\end{align*}
The per-image F1 score balances the two quantities and is given by
\[
\mathrm{F1}_i 
= 2\frac{\mathrm{P}_i\mathrm{R}_i}{\mathrm{P}_i+\mathrm{R}_i}
= \frac{2\mathrm{TP}_i}{2\mathrm{TP}_i + \mathrm{FP}_i + \mathrm{FN}_i}.
\]
Similarly, let the \textit{Intersection over Union} (IoU) measuring the overlap between the detected crack region and the ground-truth crack region be defined by 
\[
\mathrm{IoU}_i
= \frac{\mathrm{TP}_i} {\mathrm{TP}_i+\mathrm{FP}_i+\mathrm{FN}_i}.
\]
All metrics are computed per image from the same binary prediction. Table~\ref{table:metrics} reports their macro averages so that every test image contributes equally to the reported result.}
%While Precision evaluates the quality of detected crack pixels, Recall evaluates the completeness of the detection. The F1 score combines both aspects into a single measure, and IoU measures the segmentation overlap. 
Crack pixels are strongly underrepresented, leading to an imbalanced classification problem. Consequently, accuracy can be misleading, since a method may achieve high accuracy by classifying most pixels as background while failing to detect cracks reliably. For this reason, the F1 score, Precision, Recall, or IoU are more informative for evaluating crack detection quality, and we therefore do not include accuracy in our comparison of methods in Table \ref{table:metrics}. 
%\edited{Since crack pixels are strongly underrepresented, accuracy can be high even when most cracks are missed. We therefore use Precision, Recall, F1 score, and IoU, and omit accuracy from Table~\ref{table:metrics}.}
%
\ifarxiv
\begin{table}[tbp]
\centering
\begin{tabular}{lcccc}
\hline
Method & F1 & IoU & P & R \\
\hline
DeepCrack & 0.228 & 0.207 & 0.210 & 0.269 \\
fine-tuned DeepCrack & 0.912 & 0.852 & 0.852 & \textbf{0.999} \\
Masked AT & 0.872 & 0.859 & 0.862 & 0.995 \\
GenAT $\Reg_{\text{PReg}}$ & \textbf{0.973} & \textbf{0.963} & \textbf{0.965} & 0.998 \\
GenAT & 0.967 & 0.953 & 0.954 & \textbf{0.999} \\
\hline
\end{tabular}
%\caption{Quantitative evaluation results and comparison of crack detection performance metrics.}
\caption{Quantitative evaluation and comparison of the methods on $200$ synthetic test image--mask pairs. All entries are macro averages of the per-image metrics.}
\label{table:metrics}
\end{table}
\fi

\subsection{Regularization parameters and results}
\begin{figure*}[t!]
\centering
\ifarxiv
\begin{subfigure}{0.48\linewidth}
    \includegraphics[width=\linewidth]{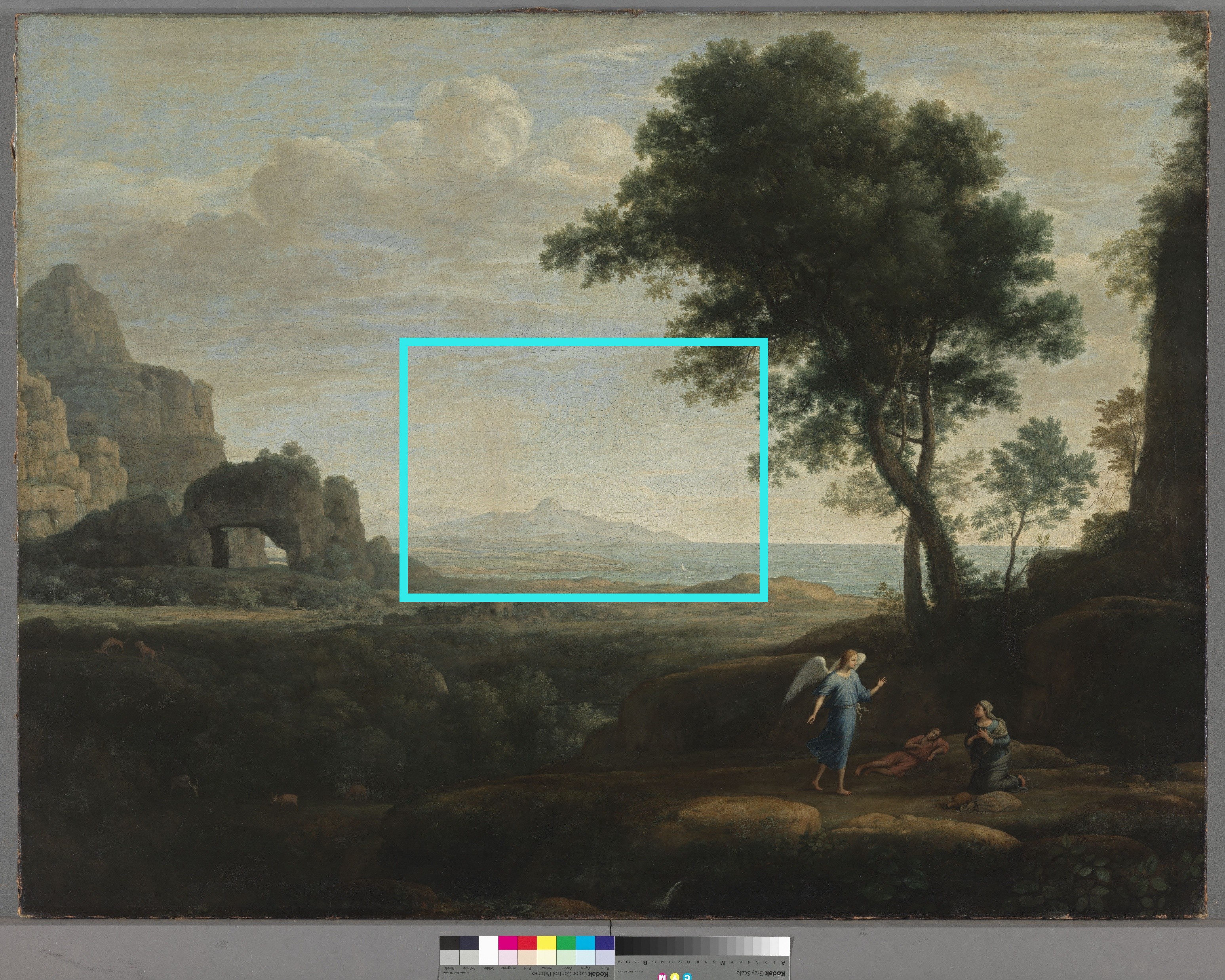}
    \caption{}
    \label{fig:Lorain_a}
\end{subfigure}\\[0.4em]
\begin{subfigure}{0.48\linewidth}
    \includegraphics[width=\linewidth]{figures/LORAIN_Wueste.jpg}
    \caption{}
    \label{fig:Lorain_b}
\end{subfigure}
\hfill
\begin{subfigure}{0.48\linewidth}
    \includegraphics[width=\linewidth]{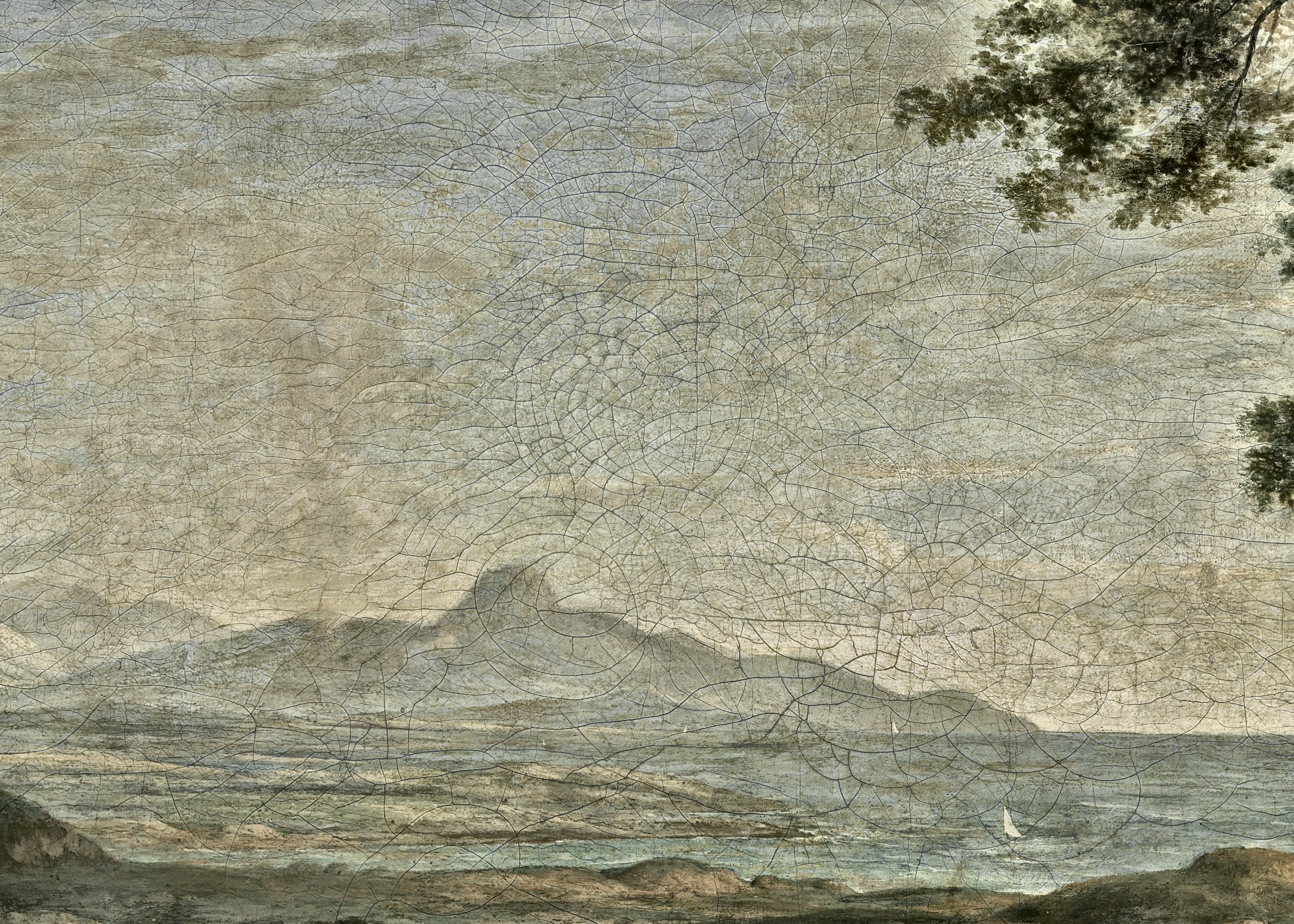}
    \caption{}
    \label{fig:Lorain_c}
\end{subfigure}\\[0.4em]
\begin{subfigure}{0.48\linewidth}
    \includegraphics[width=\linewidth]{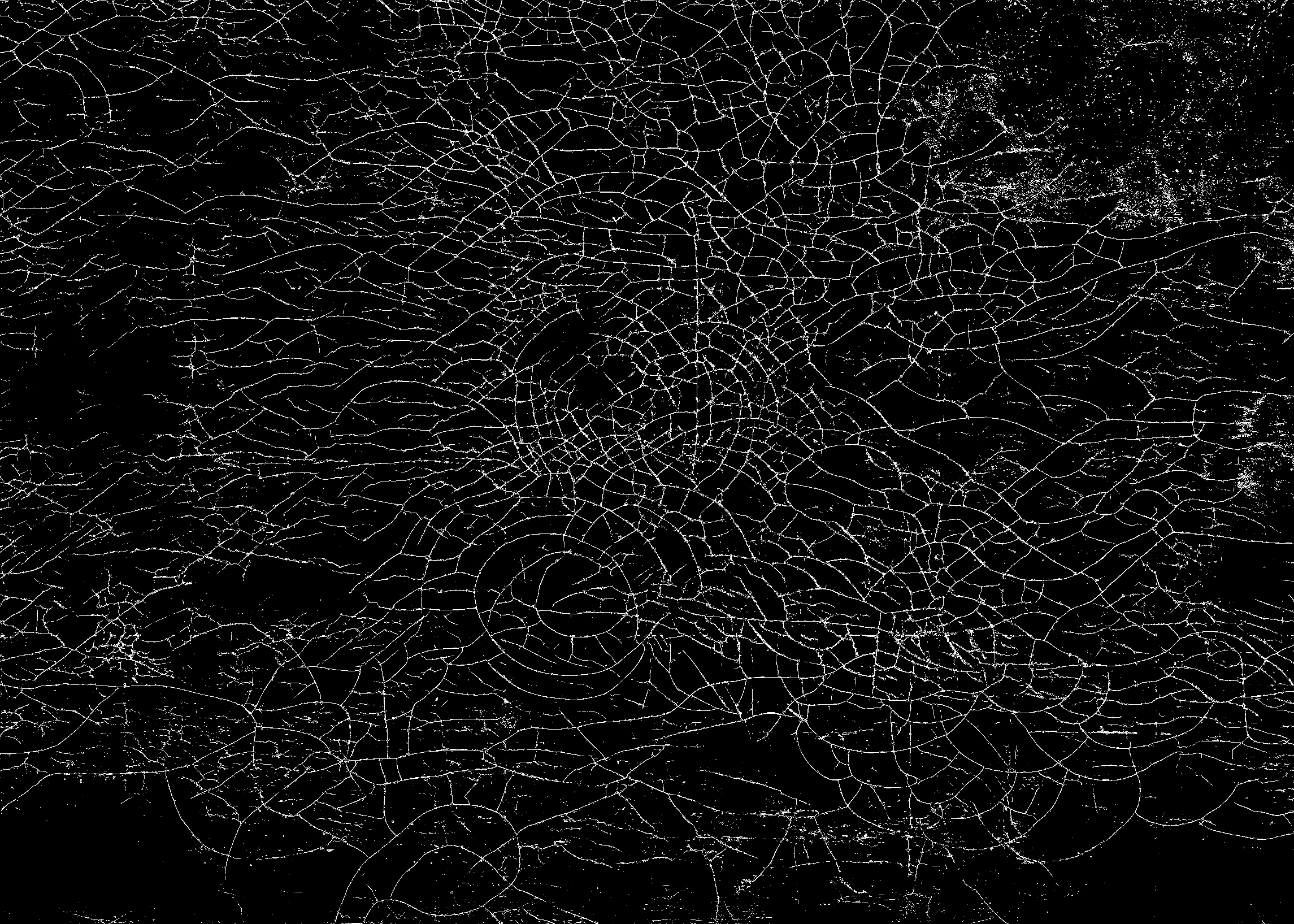}
    \caption{}
    \label{fig:Lorain_d}
\end{subfigure}
\hfill
\begin{subfigure}{0.48\linewidth}
    \includegraphics[width=\linewidth]{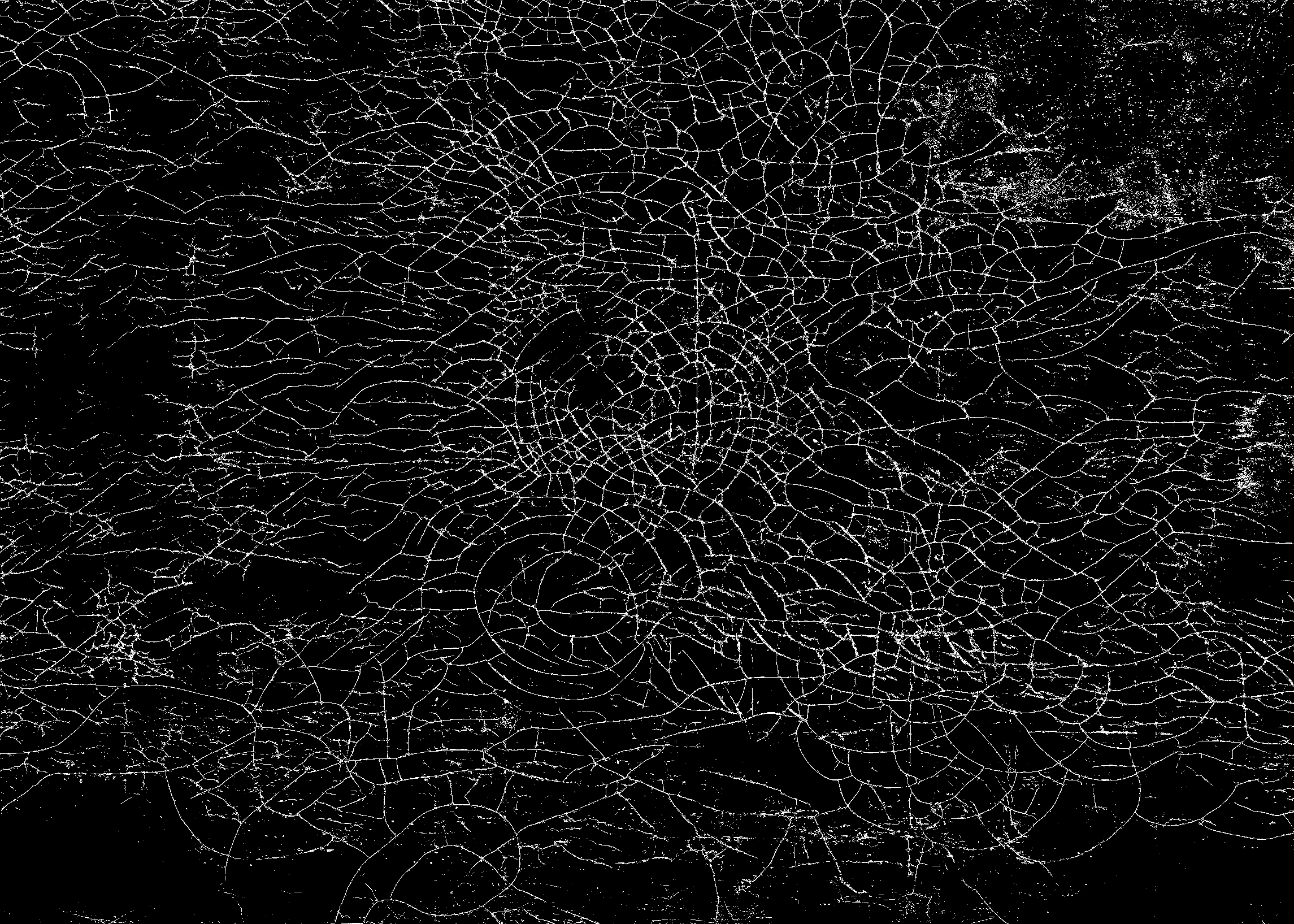}
    \caption{}
    \label{fig:Lorain_e}
\end{subfigure}
\else
\begin{subfigure}{0.3\linewidth} %0.352
    \includegraphics[width=\linewidth]{figures/LORAIN_Wueste_Ausschnitt.jpg}
    \caption{}
    \label{fig:Lorain_a}
\end{subfigure}\\[0.3em]
\begin{subfigure}{0.43\linewidth}
    \includegraphics[width=\linewidth]{figures/LORAIN_Wueste.jpg}
    \caption{}
    \label{fig:Lorain_b}
\end{subfigure}
\hfill
\begin{subfigure}{0.43\linewidth}
    \includegraphics[width=\linewidth]{figures/LORAIN_Wueste_clahe0p010.jpg}
    \caption{}
    \label{fig:Lorain_c}
\end{subfigure}\\[0.3em]
\begin{subfigure}{0.43\linewidth}
    \includegraphics[width=\linewidth]{figures/LORAIN_Wueste_FT4_ufdcfix_gT_lab_clahe_clahe0p010_kernel192_2mu_otsu_crack_bin_iter800.png}
    \caption{}
    \label{fig:Lorain_d}
\end{subfigure}
\hfill
\begin{subfigure}{0.43\linewidth}
    \includegraphics[width=\linewidth]{figures/LORAIN_Wueste_FT5_ufdcfix_gF_lab_clahe_clahe0p010_kernel192_2mu_otsu_crack_bin_iter800.png}
    \caption{}
    \label{fig:Lorain_e}
\end{subfigure}
\fi
\caption{Qualitative crack detection result of the proposed GenAT framework on a real digitized painting with structural damage. (a) Claude Lorrain (Claude Gellée), Hagar und Ismael in der Wüste, 1668, Bayerische Staatsgemäldesammlungen – Alte Pinakothek München; (b) cropped painting detail; (c) CLAHE enhanced painting detail; (d) result of GenAT with $\Reg_{\text{PReg}}$ after Otsu thresholding, and (e) result of GenAT without $\Reg_{\text{PReg}}$ after Otsu thresholding.}
\label{fig:Lorain}
\end{figure*}

\ifarxiv
\begin{figure*}[tbp]
\centering
\begin{subfigure}{0.16\linewidth}
    \includegraphics[width=\linewidth]{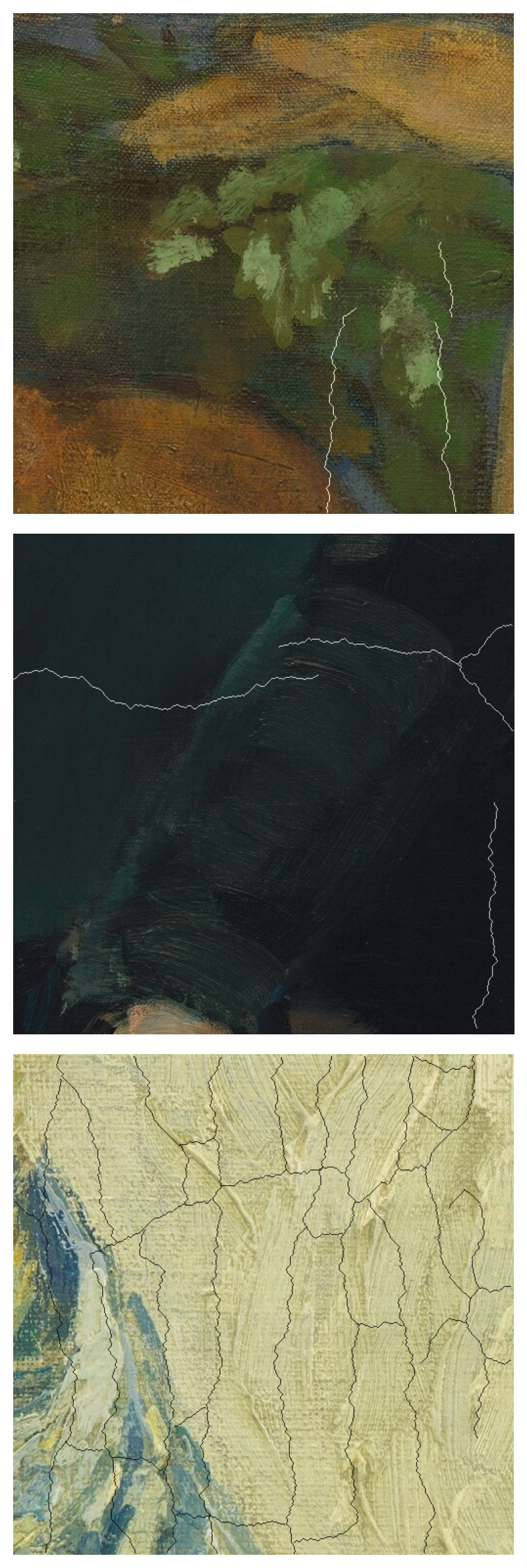}
    \caption{}
    \label{fig:comp_a}
\end{subfigure}
\hfill
\begin{subfigure}{0.16\linewidth}
    \includegraphics[width=\linewidth]{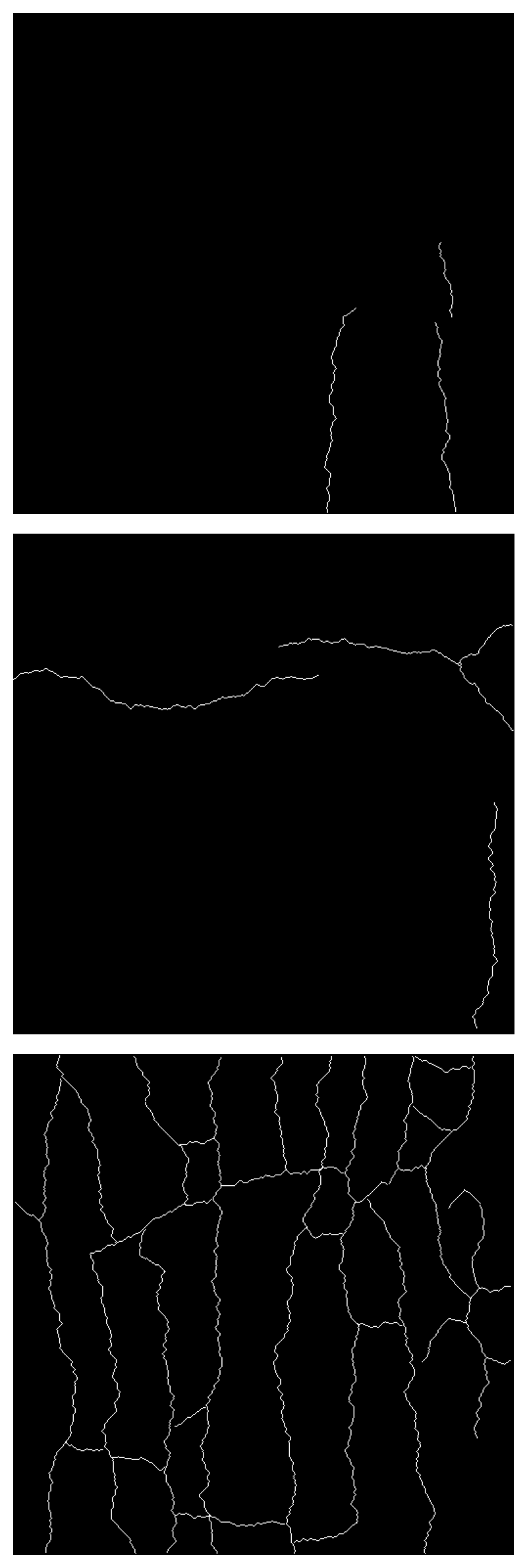}
    \caption{}
    \label{fig:comp_b}
\end{subfigure}
\hfill
\begin{subfigure}{0.16\linewidth}
\begin{tikzpicture}%
\node[anchor=south west,inner sep=0] at (0,0) 
{\includegraphics[width=\linewidth]{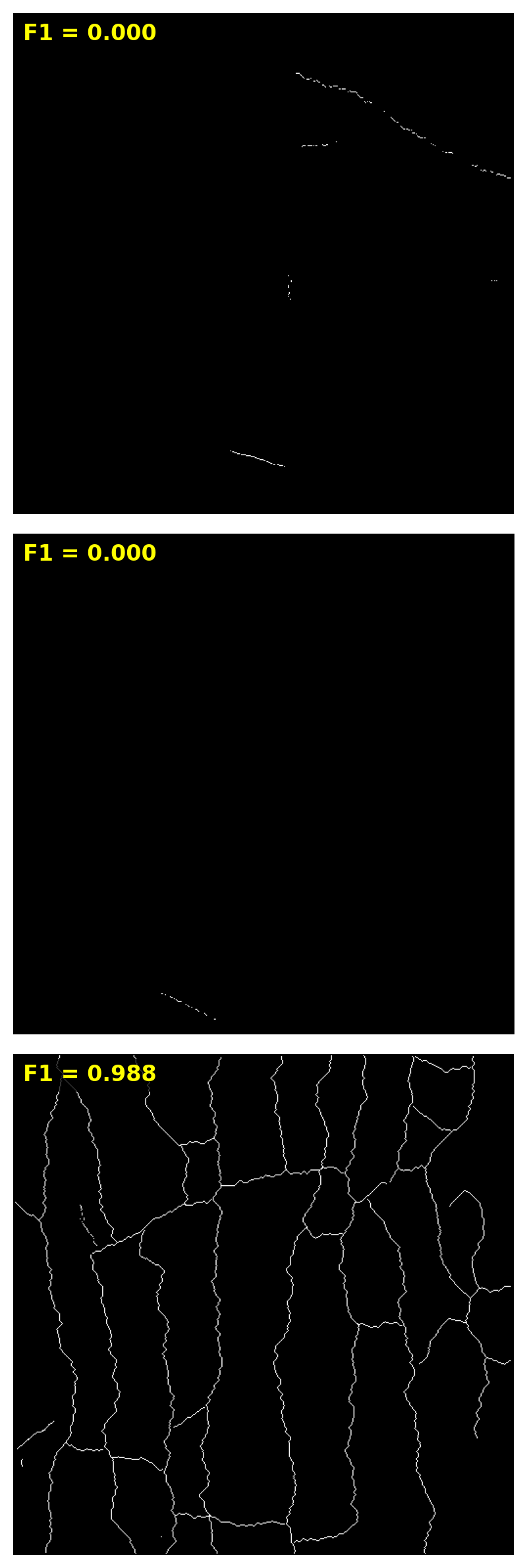}};
\draw[red,thick] (.43,.8) rectangle (0.08,.4);
\draw[red,thick] (.3,1.5) rectangle (.7,1.9);
\end{tikzpicture}
\caption{}
\label{fig:comp_c}
\end{subfigure}
\hfill
\begin{subfigure}{0.16\linewidth}
\begin{tikzpicture}%
\node[anchor=south west,inner sep=0] at (0,0) {\includegraphics[width=\linewidth]{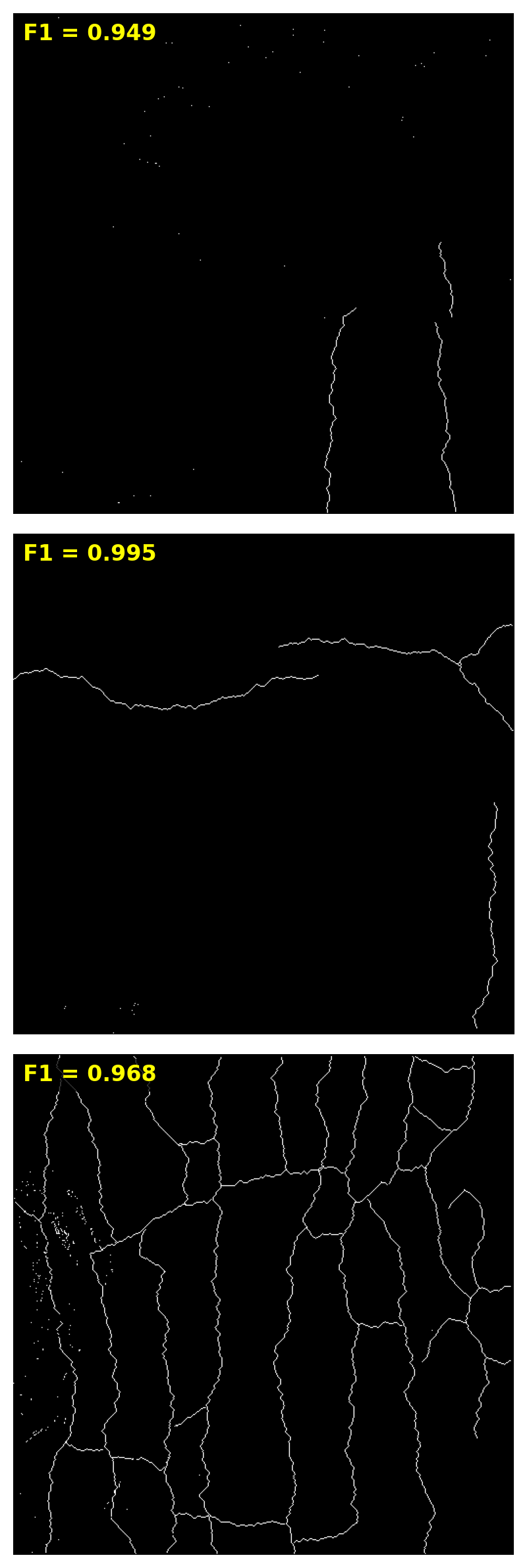}};
\draw[red,thick] (.6,.7) rectangle (.1,2.);
\end{tikzpicture}
\caption{}
\label{fig:comp_d}
\end{subfigure}
\hfill
\begin{subfigure}{0.16\linewidth}
\begin{tikzpicture}
\node[anchor=south west,inner sep=0] at (0,0) 
{\includegraphics[width=\linewidth]{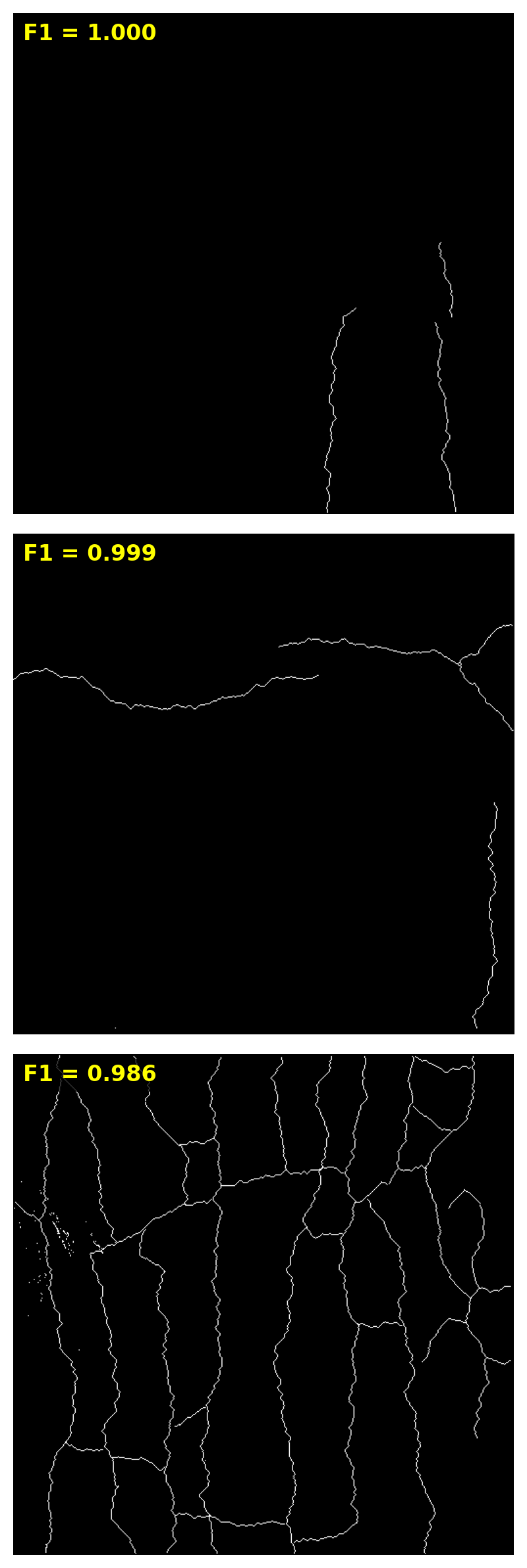}};
\draw[red,thick] (.6,1.4) rectangle (.1,2.);
\end{tikzpicture}
\caption{}
\label{fig:comp_e}
\end{subfigure}
\hfill
\begin{subfigure}{0.16\linewidth}
\begin{tikzpicture}
\node[anchor=south west,inner sep=0] at (0,0) 
{\includegraphics[width=\linewidth]{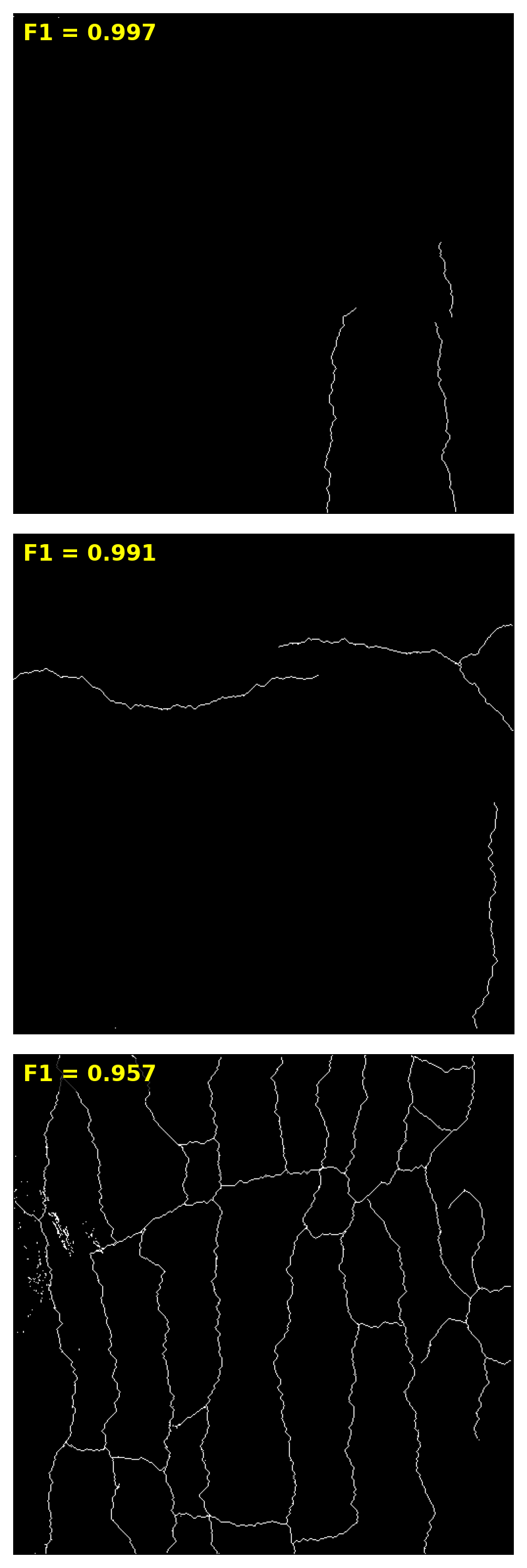}};
\draw[red,thick] (.6,1.4) rectangle (.1,2.);
\end{tikzpicture}
\caption{}
\label{fig:comp_f}
\end{subfigure}
\caption{Qualitative results of DeepCrack, simple AT with masked data fidelity term (Masked AT), and the proposed GenAT crack detection framework with and without the mixed gradient term $\Reg_{\text{PReg}}$ on synthetic test data. (a) original input image, courtesy NGA, (b) ground-truth crack mask, (c) DeepCrack result, (d) result from Masked AT, (e) result of GenAT with $\Reg_{\text{PReg}}$ after Otsu thresholding, and (f) result of GenAT without $\Reg_{\text{PReg}}$ after Otsu thresholding. Harder to distinguish artifacts are highlighted with a red box.}
\label{fig:comparison_pics}
\end{figure*}
\fi
The outcome of the iterative reconstruction based on the variational functional heavily relies on its regularization parameters. We therefore perform a grid search over the parameters $\lambda_{\text{PReg}},~\lambda_{\text{CReg}},~\lambda_{\text{CP}}$ as well as the weight $\varepsilon$ that appears in the definition of $\Reg_\text{CReg}^\varepsilon$. 
\edited{For each configuration, we calculate the macro-averaged validation F1 score on a subset of five images from a held-out validation set containing $208$ synthetic image patches and masks and select the configuration attaining the largest value. After parameter selection, the inference procedure is fixed and evaluated on a separate test set of $200$ image-mask pairs.} 
The selected parameters are %
\[ 
\lambda_{\text{PReg}}=1, \, \lambda_{\text{CReg}}=0.1, \, \lambda_{\text{CP}}=0.5 \text{ and } \varepsilon=0.005
\]%
\edited{for the model including the term $\Reg_{\text{PReg}}$ defined in (\ref{GradientTerm}) and %
% \[ 
% \lambda_{\text{CReg}}=0.5, \, \lambda_{\text{CP}}=0.5 \text{ and } \varepsilon=0.05
% \]
% for the model without it ($\lambda_{\text{PReg}}=0$).
\[
\lambda_{\text{PReg}}=0,\ \lambda_{\text{CReg}}=0.5,\ \lambda_{\text{CP}}=0.5 \text{ and } \varepsilon=0.05,
\]
otherwise.} %
For the simple Ambrosio--Tortorelli approach with masked data fidelity term, we use $700$ optimization iterations. For the proposed GenAT framework, we use $300$ iterations for experiments on the synthetic dataset and $800$ iterations for detection on real-world digitized paintings.
\ifarxiv
\else
\begin{table}[tbp]
\centering
\begin{tabular}{lcccc}
\hline
Method & F1 & IoU & P & R \\
\hline
DeepCrack & 0.228 & 0.207 & 0.210 & 0.269 \\
Fine-tuned DeepCrack & 0.912 & 0.852 & 0.852 & \textbf{0.999} \\
Masked AT & 0.872 & 0.859 & 0.862 & 0.995 \\
GenAT $\Reg_{\text{PReg}}$ & \textbf{0.973} & \textbf{0.963} & \textbf{0.965} & 0.998 \\
GenAT & 0.967 & 0.953 & 0.954 & \textbf{0.999} \\
\hline
\end{tabular}
%\caption{Quantitative evaluation results and comparison of crack detection performance metrics.}
\caption{\edited{Quantitative evaluation and comparison of the methods on $200$ synthetic test image-mask pairs. All entries are macro averages of the per-image metrics.}}
\label{table:metrics}
\end{table}
\fi
The final crack map is computed from the optimized variable $v$ as $\C_{\text{pred}} = 1 - v$, then a binary pixel-wise crack map is obtained by thresholding via Otsu's method \cite{otsu}. Otsu's method automatically determines a threshold that best separates crack and non-crack pixels by maximizing the difference between the two groups while minimizing the intensity variation within each group. \edited{For the original and fine-tuned DeepCrack models, the predicted crack probabilities are binarized using a fixed threshold of $0.5$.} Using the specified parameters, the performance of the proposed method \edited{on the test set split of $200$ images} is reported in Table~\ref{table:metrics} and compared with the original DeepCrack model, its fine-tuned version, and the AT approach using a weighted data-fidelity term (\emph{Masked AT}). Qualitative crack detection results on the synthetic dataset are shown in Figure \ref{fig:comparison_pics}, comparing DeepCrack, Masked AT, and the proposed GenAT framework, both with and without the mixed gradient regularization term $\Reg_{\mathrm{PReg}}$. The quantitative results in Table \ref{table:metrics} as well as the results in Figure \ref{fig:comp_c} indicate that the original DeepCrack model does not transfer well to art-specific data containing complex painting structures. This can be attributed to the domain gap between pavement crack images and digitized paintings and the fact that DeepCrack can mainly detect dark cracks. Fine-tuning DeepCrack on art-related data with varying crack appearance substantially improves its performance, making it a viable crack prior for the proposed framework. Nevertheless, the model still struggles in the presence of crack-like brushstrokes and strong intensity variations in the background, which are occasionally misclassified as cracks. According to the results in Table \ref{table:metrics}, the Masked AT model alone is likewise insufficient for high-quality crack detection. While it captures a large proportion of the true crack pixels, it also produces a considerable number of FP. This effect is visible in Figure \ref{fig:comp_d}, particularly in the red-marked regions of the last row. Combining the Masked AT formulation with the fine-tuned generative VQGAN prior for the smooth painting background significantly improves the crack detection performance and yields the best overall quantitative results. On the synthetic test dataset, the variant including the mixed gradient regularization term (GenAT $\Reg_{\mathrm{PReg}}$) achieves the highest performance. For real-world cracked paintings, however, the advantage of including $\Reg_{\mathrm{PReg}}$ is less pronounced, and no clear superiority can be observed from the qualitative results alone. Figures \ref{fig:teaser}, \ref{fig:Zecher}, and \ref{fig:Lorain} illustrate the performance of the proposed approach on real-world digitized paintings. \edited{Figure \ref{fig:Zecher} shows \textit{Der Zecher} by Hendrick ter Brugghen from $1627$, which measures $71.3 \times 60$ cm. Figure \ref{fig:Lorain} shows \textit{Hagar und Ismael in der Wüste} by Claude Lorrain from $1668$, measuring $106.4 \times 140$ cm. Both paintings were digitized using a Hasselblad H4D 60 camera with 60 megapixels at a resolution of $8377 \times 6616$ pixels. An Elinchrom flash system provided uniform illumination in a predominantly neutral gray environment to minimize stray light and color casts while color fidelity was ensured through white balancing against a daylight reference and comparison of a profiled print with the original painting.} 
\ifarxiv
\begin{figure*}[tbp]
\centering
\begin{subfigure}{0.48\linewidth}
    \centering
    \includegraphics[height=0.243\textheight]
    {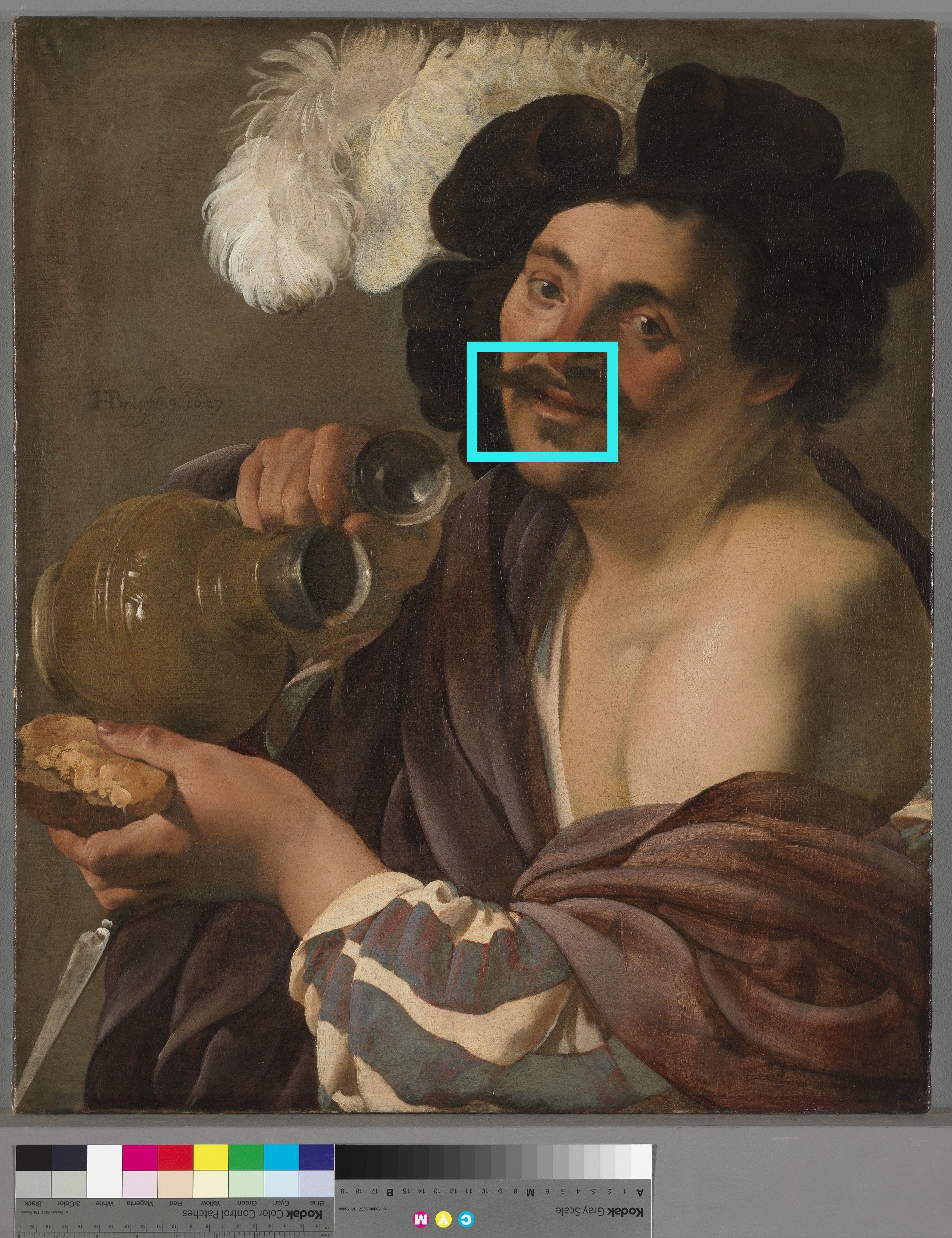}
    \caption{}
    \label{fig:Zecher_a}
\end{subfigure}
\hfill
\begin{subfigure}{0.48\linewidth}
    \centering
    \includegraphics[width=\linewidth]{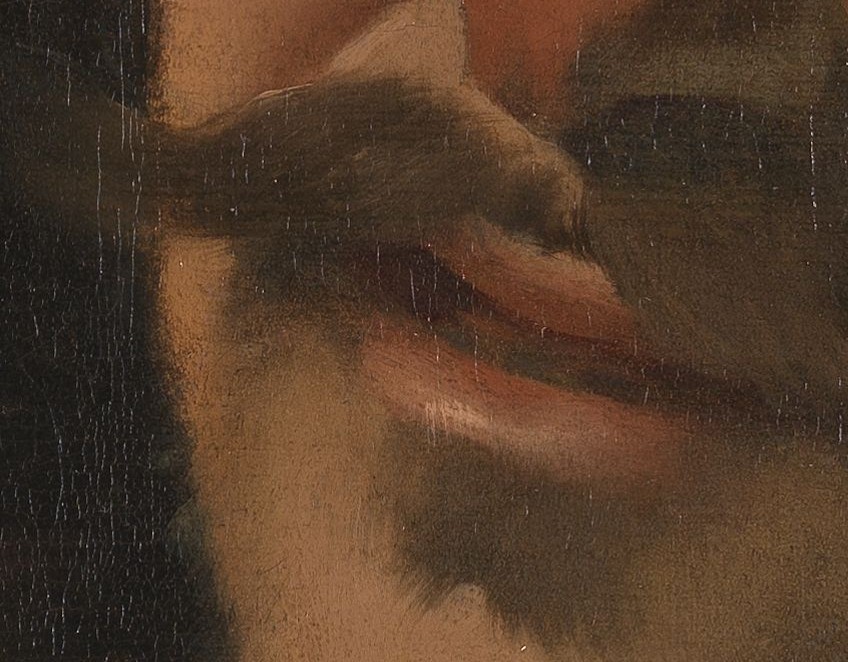}
    \caption{}
    \label{fig:Zecher_b}
\end{subfigure}\\[0.4em]
\begin{subfigure}{0.48\linewidth}
    \includegraphics[width=\linewidth]{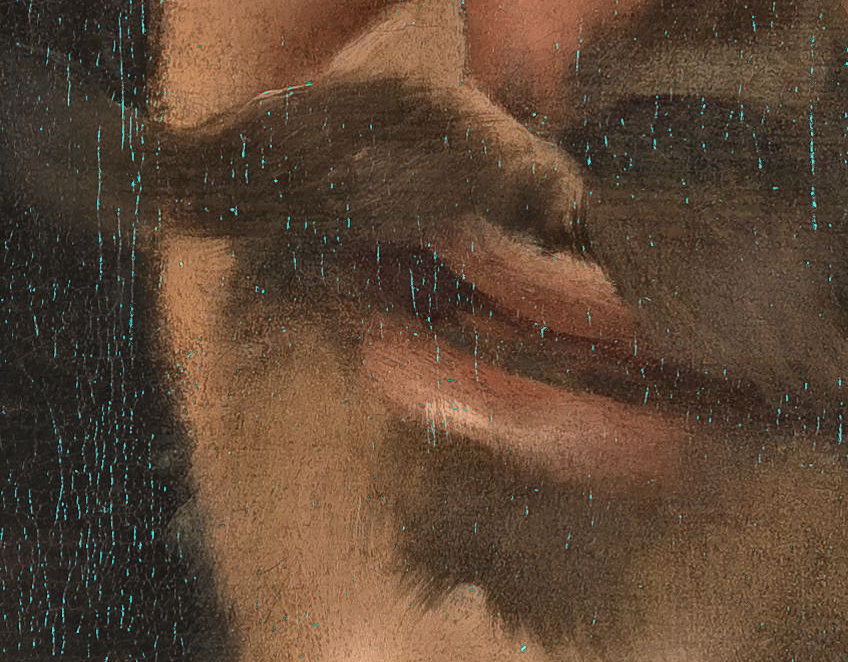}
    \caption{}
    \label{fig:Zecher_c}
\end{subfigure}
\hfill
\begin{subfigure}{0.48\linewidth}
    \includegraphics[width=\linewidth]{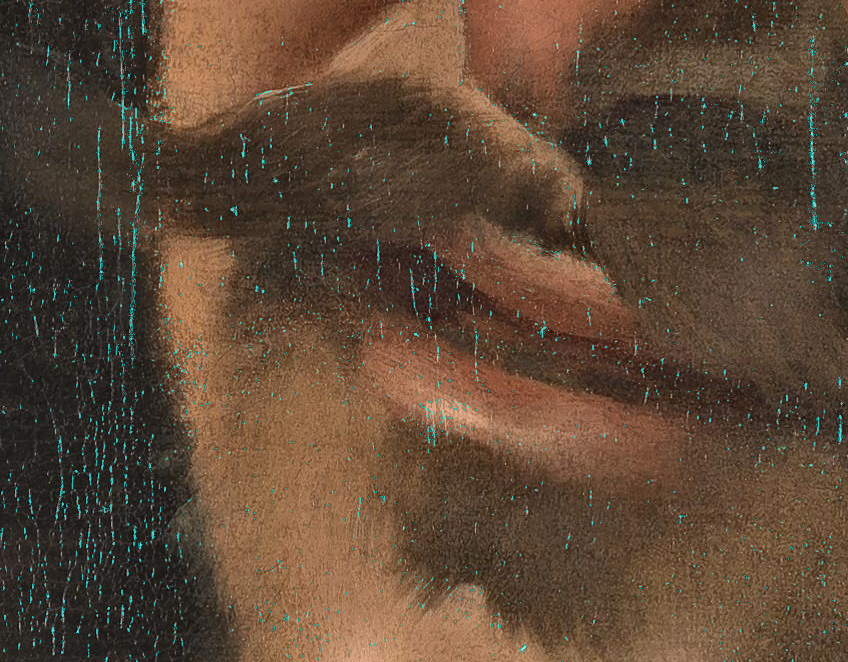}
    \caption{}
    \label{fig:Zecher_d}
\end{subfigure}
\caption{Qualitative crack detection result of the proposed GenAT framework on a real digitized painting with structural damage. (a) Hendrick ter Brugghen, Der Zecher, 1627, Bayerische Staatsgemäldesammlungen - Alte Pinakothek München; (b) cropped painting detail; (c) result of GenAT with $\Reg_{\text{PReg}}$ and (d) without $\Reg_{\text{PReg}}$, where in both cases the Otsu-thresholded binary crack map is overlaid in cyan on the painting detail.}
\label{fig:Zecher}
\end{figure*}
\else
\begin{figure*}[ht]%[tbp]
\centering
\begin{subfigure}{0.43\linewidth}
    \centering
    \includegraphics[height=0.266\textheight] %0.297
    {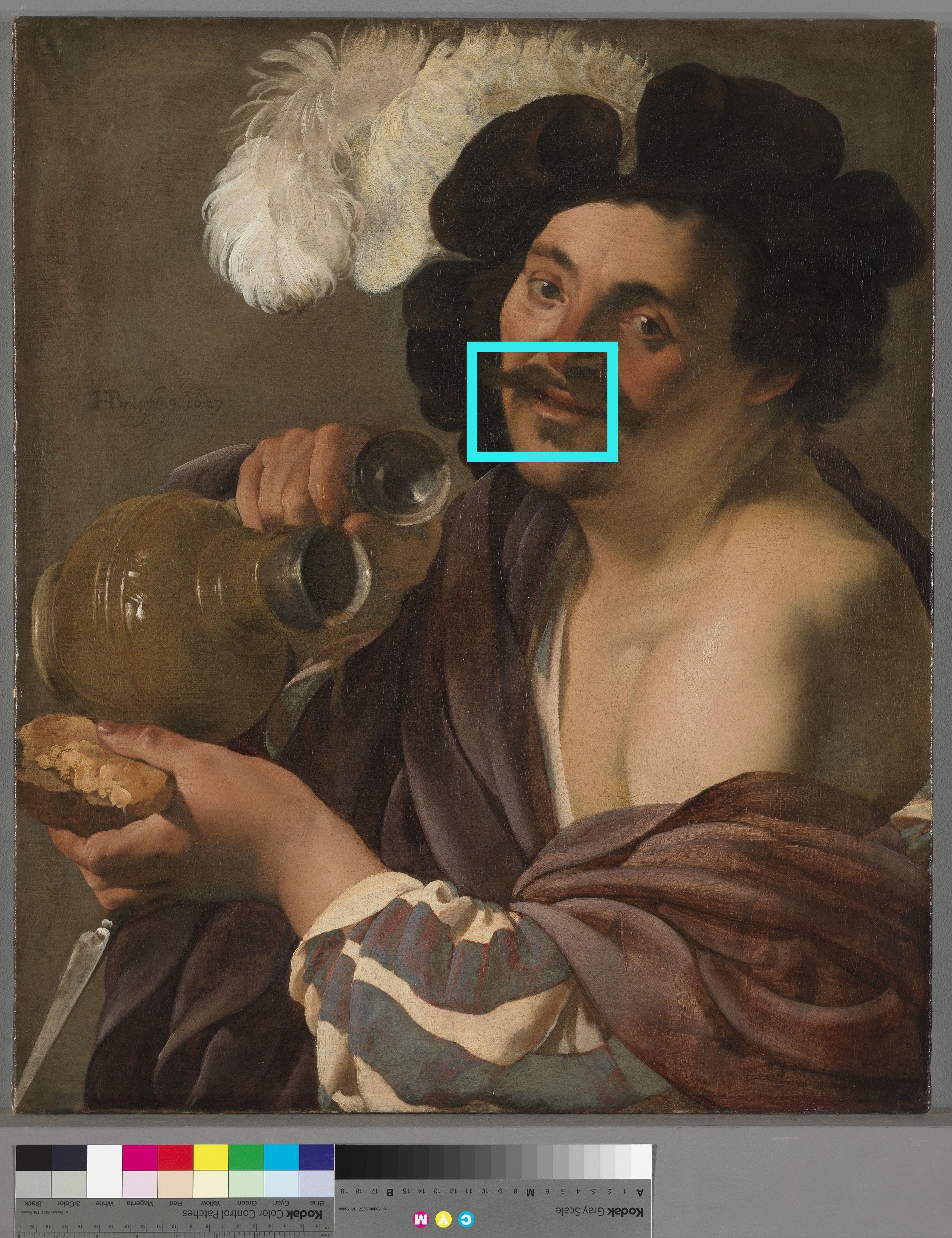}
    \caption{}
    \label{fig:Zecher_a}
\end{subfigure}
\hfill
\begin{subfigure}{0.43\linewidth}
    \centering
    \includegraphics[width=\linewidth]{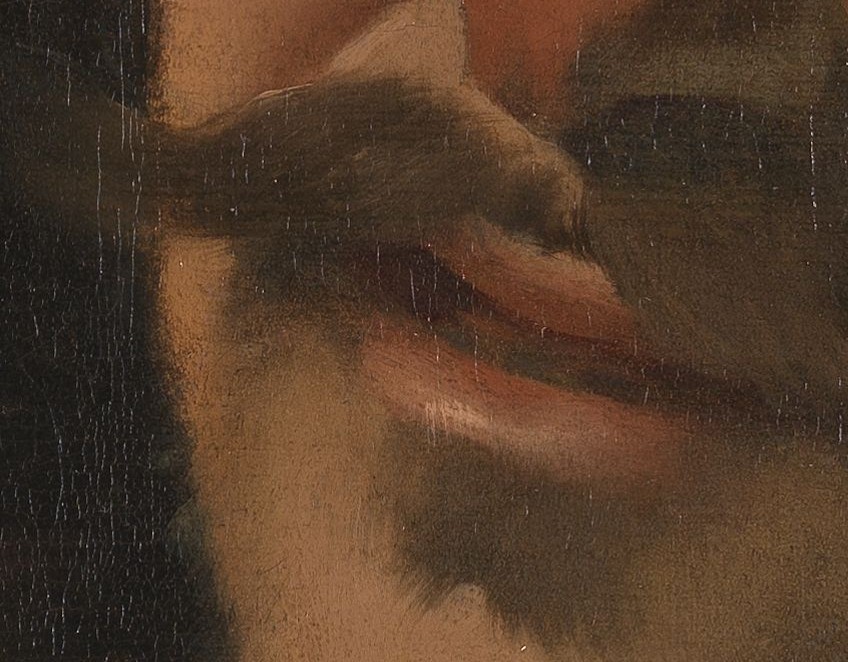}
    \caption{}
    \label{fig:Zecher_b}
\end{subfigure}\\[0.3em]
\begin{subfigure}{0.43\linewidth}
    \includegraphics[width=\linewidth]{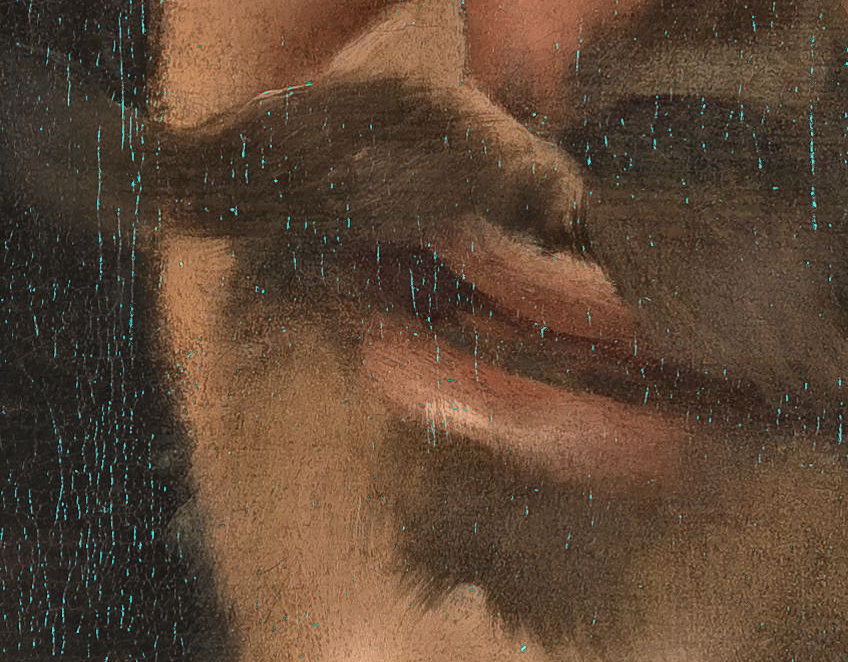}
    \caption{}
    \label{fig:Zecher_c}
\end{subfigure}
\hfill
\begin{subfigure}{0.43\linewidth}
    \includegraphics[width=\linewidth]{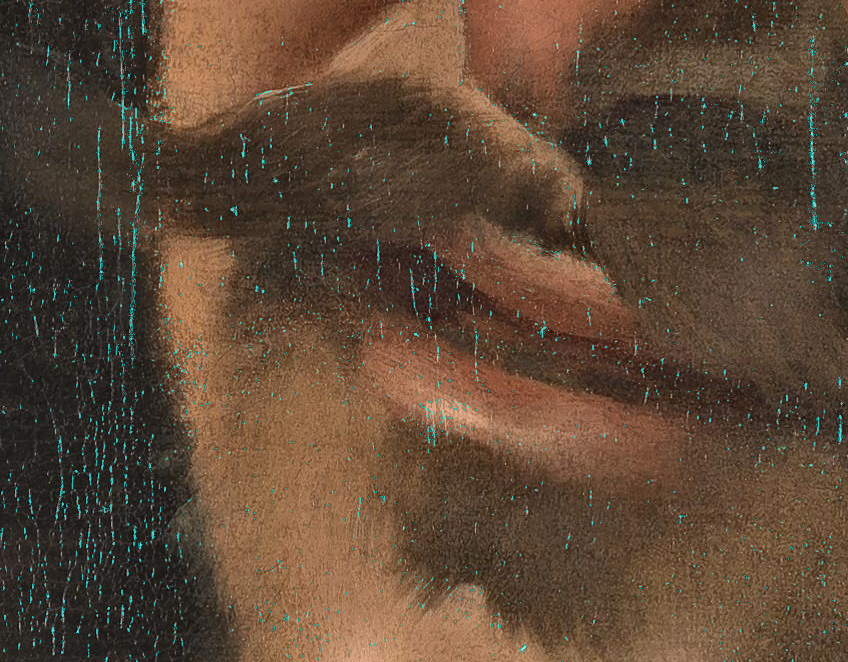}
    \caption{}
    \label{fig:Zecher_d}
\end{subfigure}
\caption{Qualitative crack detection result of the proposed GenAT framework on a real digitized painting with structural damage. (a) Hendrick ter Brugghen, Der Zecher, 1627, Bayerische Staatsgemäldesammlungen - Alte Pinakothek München; (b) cropped painting detail; (c) result of GenAT with $\Reg_{\text{PReg}}$ and (d) without $\Reg_{\text{PReg}}$, where in both cases the Otsu-thresholded binary crack map is overlaid in cyan.}
\label{fig:Zecher}
\end{figure*}
\fi%
To improve the visibility of fine cracks, especially in dark or low-contrast regions, an optional image enhancement step is applied to the real-world digitized paintings before crack detection with our approach when needed. Instead of modifying the colors directly, the method enhances only the brightness information of the image in the perceptually motivated LAB color space. Local contrast is increased using contrast limited adaptive histogram equalization (CLAHE) \cite{clahe1994}, which enhances weak crack structures while avoiding excessive global contrast changes. To suppress noise and texture artifacts introduced by this enhancement, an optional mild edge-preserving denoising filter can additionally be applied. This preprocessing makes faint cracks more distinguishable without substantially altering the visual appearance of the painting. For the painting patch from the painting by Claude Lorrain, the CLAHE-enhanced version is displayed in Figure \ref{fig:Lorain_c}. For the Zecher painting patch, only mild CLAHE preprocessing was applied, since the appearing cracks are very thin and stronger local contrast enhancement tended to amplify the crack responses excessively, resulting in wider and less localized crack structures and an increased number of false detections. Furthermore, for such larger or high-resolution image files, crack maps are computed on overlapping patches using local thresholding. Crack maps are averaged in overlapping regions, while binary crack detections are merged using a logical OR to preserve thin structures. %
\edited{Across the two paintings, the effect of $\Reg_{\mathrm{PReg}}$ is image-dependent. For \textit{Der Zecher}, Figure \ref{fig:Zecher} suggests that omitting $\Reg_{\mathrm{PReg}}$ (Figure \ref{fig:Zecher_d}) leads to the detection of more cracks, but at the cost of an increased number of false positives. In contrast, for the Lorrain painting, including $\Reg_{\mathrm{PReg}}$ recovers additional crack structures, as visible in the lower-left region of Figures~\ref{fig:Lorain_d} and~\ref{fig:Lorain_e}. 
Nevertheless, the non-thresholded crack map shown in Figure \ref{fig:teaser} indicates that several subtle crack structures that are absent from the Otsu-thresholded crack maps in Figures \ref{fig:Lorain_d} and \ref{fig:Lorain_e} are detected by the model, but with lower confidence. Here brighter pixels correspond to a higher estimated probability of belonging to a crack. Faint structures in the left-central region remain visible as low-intensity responses in the non-thresholded crack map, but are removed by Otsu thresholding and therefore do not appear in the binary maps in Figures~\ref{fig:Lorain_d} and~\ref{fig:Lorain_e}.} 
\Doerner{The results on the real paintings were further assessed by conservation experts at the Doerner Institut through comparison with the original works. Overall, they concluded that the proposed crack detection performed well in most regions and largely agreed with their observations of the paintings. For \textit{Der Zecher}, it was confirmed that many of the detected pixels corresponded to actual craquelure, although the detections did not form a continuous crack network and fine cracks were frequently missed, particularly in the dark region on the left side of the painting detail in Figures~\ref{fig:Zecher_c} and ~\ref{fig:Zecher_d}. Some undetected cracks could only be recognized through direct examination of the physical painting under illumination from different angles and may therefore not be clearly represented in a single digitized photograph acquired under fixed illumination. For the Lorrain painting, a large proportion of the craquelure was recovered as a connected network closely resembling the actual pattern, particularly in the bright sky regions. However, Figures \ref{fig:Lorain_d} and \ref{fig:Lorain_e} show that high-contrast painted details, for example in the tree or along the horizon, were falsely detected as cracks. 
These findings indicate that the resulting maps can support screening and documentation, but should be interpreted and verified by conservation experts.}

%%%%%%%%%%%%%%%%%%%%%%%%%%%%%%%%%%%%%%%%%%%%%%%%%%%%%%%%%%%%%%%%%%%%%%%%
\section{Summary and Conclusions}
\noindent In this work, we introduced a variational inverse problem formulation for crack detection in digitized paintings that combines classical image regularization with modern powerful learned priors. Crack detection is modeled as the joint recovery of a crack-free painting and a crack map, where the background is represented by a pretrained and fine-tuned generative model, while crack structures are recovered via a Mumford--Shah-type regularization guided by an additional crack prior. The proposed GenAT approach allows pixel-level crack localization without relying on image pre- or postprocessing, apart from optional contrast enhancement and a final thresholding step. 
\edited{The theoretical framework introduced in Section \ref{sec:theory} is not tied to a specific choice of crack prior and permits an unsupervised implementation. However, the evaluated approach is weakly supervised through the DeepCrack prior and the parameter selection on labeled synthetic data.} 
Synthetic data generation using realistic crack patterns enables quantitative evaluation in the absence of annotated artwork datasets. On the synthetic test set, the numerical experiments demonstrate that most crack structures are correctly identified while relatively few false positives are introduced, despite the presence of complex background textures. This indicates that structural analysis in CH imaging and specifically crack detection in digitized artworks can be effectively addressed as a regularized inverse problem with learned priors, offering an alternative to fully supervised approaches. Deep learning-guided variational models with generative image representations thus constitute a promising direction for crack detection and, more broadly, structural analysis in CH imaging. \edited{Feedback from conservation experts indicated that the detections largely matched the craquelure observed in the original paintings, while confirming that fine cracks may be missed and high-contrast painted details may produce false positives. Given the weakly supervised setting and the absence of pixel-wise annotations for real paintings, such errors are not unexpected.} One possible cause is that the synthetic dataset may not fully capture the diversity of natural aging crack patterns. Additionally, hyperparameters working best on the synthetic dataset may not be optimal for real-world images. 
%Future work could investigate strategies to improve generalization. 
\edited{Diverse painting backgrounds in the synthetic test set provide initial evidence of robustness to varied visual content, although broader experiments on real paintings beyond the two works from the seventeenth century are needed to assess generalization across historical periods, painting materials, and degradation conditions. When available, complementary multispectral or multi-light imaging could improve the visibility of faint cracks and their distinction from painted details, with multi-light imaging particularly relevant for cracks visible only under changing illumination. GenAT could be extended to couple these complementary observations within a joint optimization that yields a single shared crack map and coupled generated background reconstructions specific to each modality. Their complementary information may improve detection performance, although integrating non-RGB data would require adapting the current components. Moreover, future work could replace DeepCrack with an unsupervised prior, such as a crack generator trained on unlabeled patterns, avoiding annotated or synthetic training data.}
%Future work could explore replacing the DeepCrack prior with an unsupervised crack prior to avoid reliance on annotated or synthetic datasets for training. This could, for example, involve a crack generator trained on unlabeled crack patterns.

\section*{Acknowledgements}
\Doerner{The authors gratefully acknowledge the Doerner Institut of the Bayerische Staatsgemäldesammlungen, represented by its director, Eva Ortner, for valuable early discussions, feedback on the crack detection procedure from a conservation perspective, and the provision of high-resolution images for evaluation.} \edited{MB and SK acknowledge support from DESY (Hamburg, Germany), a member of the Helmholtz Association HGF and from the DFG, project BU 2327/19-1. %
TR acknowledges the support of the Munich Center for Machine Learning. Funded by the European Union. Views and opinions expressed are however those of the author(s) only and do not necessarily reflect those of the European Union or the European Research Council Executive Agency. Neither the European Union nor the granting authority can be held responsible for them. This project has received funding from the European Research Council (ERC) under the European Union’s Horizon Europe research and innovation programme (grant agreement No. 101198055, project acronym NEITALG).
\begin{center}
\includegraphics[height=1cm]{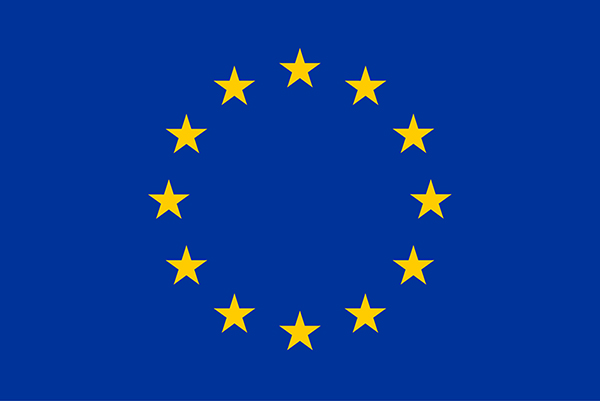}%
\hspace{1em}
\includegraphics[height=1cm]{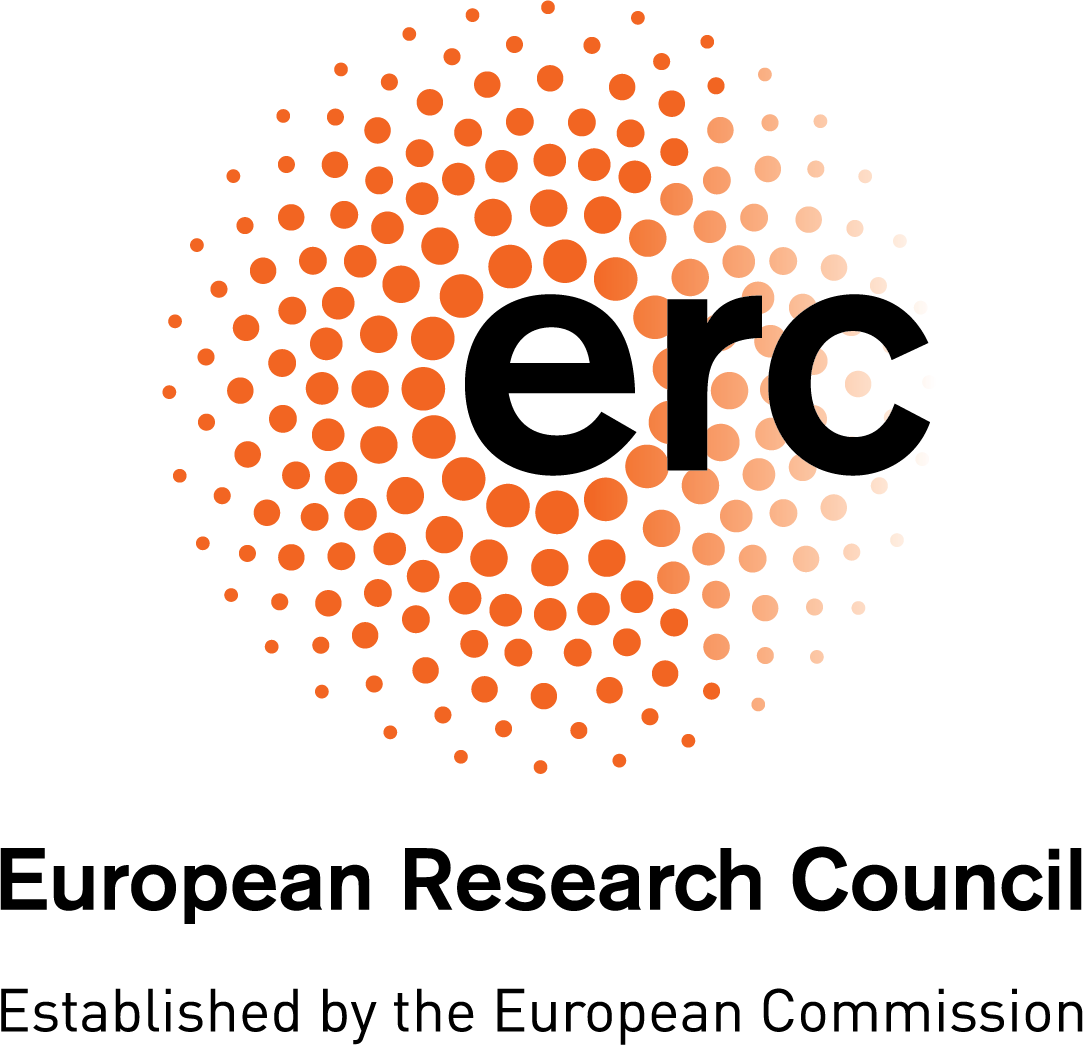}%
\end{center}
}

%-------------------------------------------------------------------------

\bibliographystyle{alpha}
\bibliography{literature.bib}

%-------------------------------------------------------------------------
%Color tables are no longer required for purely electronic publications.

%\newpage
%large figures here for printed version

\end{document}